\documentclass{article} 
\usepackage[usenames,dvipsnames]{xcolor}
\usepackage{iclr2024_conference,times}

\usepackage{array}
\usepackage{ulem}
\usepackage{amsmath}
\usepackage{amsfonts}       
\usepackage{nicefrac}       
\usepackage{microtype}      
\usepackage{hyperref}
\usepackage{url}
\usepackage{multirow}
\usepackage{xcolor}
\usepackage{float}
\usepackage{comment}
\usepackage{diagbox}
\usepackage{wrapfig}
\usepackage{graphicx}
\usepackage{algorithm}
\usepackage{algpseudocode}

\usepackage[all]{hypcap}
\usepackage{graphicx}
\usepackage{mathtools}
\usepackage{array}
\usepackage{caption}
\usepackage{subcaption}
\usepackage{ulem}
\newcolumntype{?}{!{\vrule width 1.5pt}}
\newcolumntype{:}{!{\vrule width 2pt}}
\usepackage{booktabs} 

\newcommand{\Pn}{{\textbf{ConR}}}

\long\def\authornote#1#2{%
  \leavevmode\unskip\raisebox{-3.5pt}{\rlap{\textcolor{#1}{$\scriptstyle\diamond$}}}%
  \marginpar{\raggedright\hbadness=10000
    \def\baselinestretch{0.8}\tiny
    \it \textcolor{#1}{#2}\par}}
\newif\ifcomments
\newcommand{\dave}[1]{\ifcomments\authornote{purple}{\textbf{Dave:} #1}\fi}

\title{ConR: Contrastive  Regularizer for Deep Imbalanced Regression}

\author{
  Mahsa Keramati\textsuperscript{1,2}\thanks{ Work done during an internship at Borealis AI.},
  Lili Meng\textsuperscript{1},
  R. David Evans\textsuperscript{1} \\
  \textsuperscript{1} Borealis AI,
  \textsuperscript{2} School of Computing Science, Simon Fraser University \\
    \texttt{mkeramat@sfu.ca},
    \texttt{lili.meng@gmail.com,dave.evans@borealisai.com} 
    \\
}

\iclrfinalcopy 
\begin{document}

\maketitle

\begin{abstract}

Imbalanced distributions are ubiquitous in real-world data. They create constraints on Deep Neural Networks to represent the minority labels and avoid bias towards majority labels.
The extensive body of imbalanced approaches address categorical label spaces but fail to effectively extend to regression problems where the label space is continuous. 
Local and global correlations among continuous labels provide valuable insights towards effectively modelling relationships in feature space. 
In this work, we propose \Pn{}, a contrastive regularizer that models global and local label similarities in feature space and prevents the features of minority samples from being collapsed into their majority neighbours. 
\Pn{} discerns the disagreements between the label space and feature space, and imposes a penalty on these disagreements. \Pn{} addresses the continuous nature of label space with two main strategies in a contrastive manner: incorrect proximities are penalized proportionate to the label similarities and the correct ones are encouraged to model local similarities.
\Pn{} consolidates essential considerations into a generic, easy-to-integrate, and efficient method that effectively addresses deep imbalanced regression.
Moreover, \Pn{} is orthogonal to existing approaches and smoothly extends to uni- and multi-dimensional label spaces. Our comprehensive experiments show that \Pn{} significantly boosts the performance of all the state-of-the-art methods on four large-scale deep imbalanced regression benchmarks.
Our code is publicly available in \href{https://github.com/BorealisAI/ConR}{https://github.com/BorealisAI/ConR}.

\end{abstract}

\section{Introduction}
Imbalanced data distributions, which are common in real-world contexts,  introduce substantial challenges in generalizing conventional models due to variance across minority labels and bias to the majority ones~\citep{wang2021longtailed, ranksim,buda2018systematic}.\dave{do we need to define minority and majority here?}
Although there are numerous works on learning from imbalanced data~\citep{chawla2002smote,cui2021parametric,jiang2021improving}, these studies mainly focus on categorical labels. 
Continuous labels are potentially infinite, high-dimensional and hard to bin semantically~\citep{ren2022balanced}. These characteristics impede the performance of imbalanced classification approaches on Deep Imbalanced Regression (DIR)~\citep{yang2021delving}. 

Continuous labels result in underlying local and global correlations, which yields valuable perspectives towards effective representation learning of imbalanced data~\citep{ranksim,shin2022moving}.
For instance, regression training on imbalanced data fails to model appropriate relationships for minority labels in the feature space~\citep{yang2021delving}. \citet{yang2021delving} established an empirical example of this phenomenon on the age estimation task where the learned features of under-represented samples collapse to majority features.
Several approaches tackle this issue by encouraging local dependencies ~\citep{yang2021delving,steininger2021density}.
However, these methods fail to exploit global relationships and are biased toward only learning representative features for majority samples, especially when minority examples do not have majority neighbours.
RankSim~\citep{ranksim} leverages global and local dependencies by exploiting label similarity orders in feature space. 
Yet, RankSim does not generalize to all regression tasks, as not all continuous label spaces convey order relationships. For example, for depth-map estimation from scene pictures, the complicated relationships in the high-dimensional label space are not trivially convertible to a linearly ordered feature space. 
\textbf{Given the importance of the correspondence between the label space and feature space for imbalanced regression, can we effectively transfer inter-label relationships, regardless of their complexity, to the feature space?}

\begin{wrapfigure}{r}{0.5\textwidth}
\centering
    \includegraphics[width=0.5\textwidth]{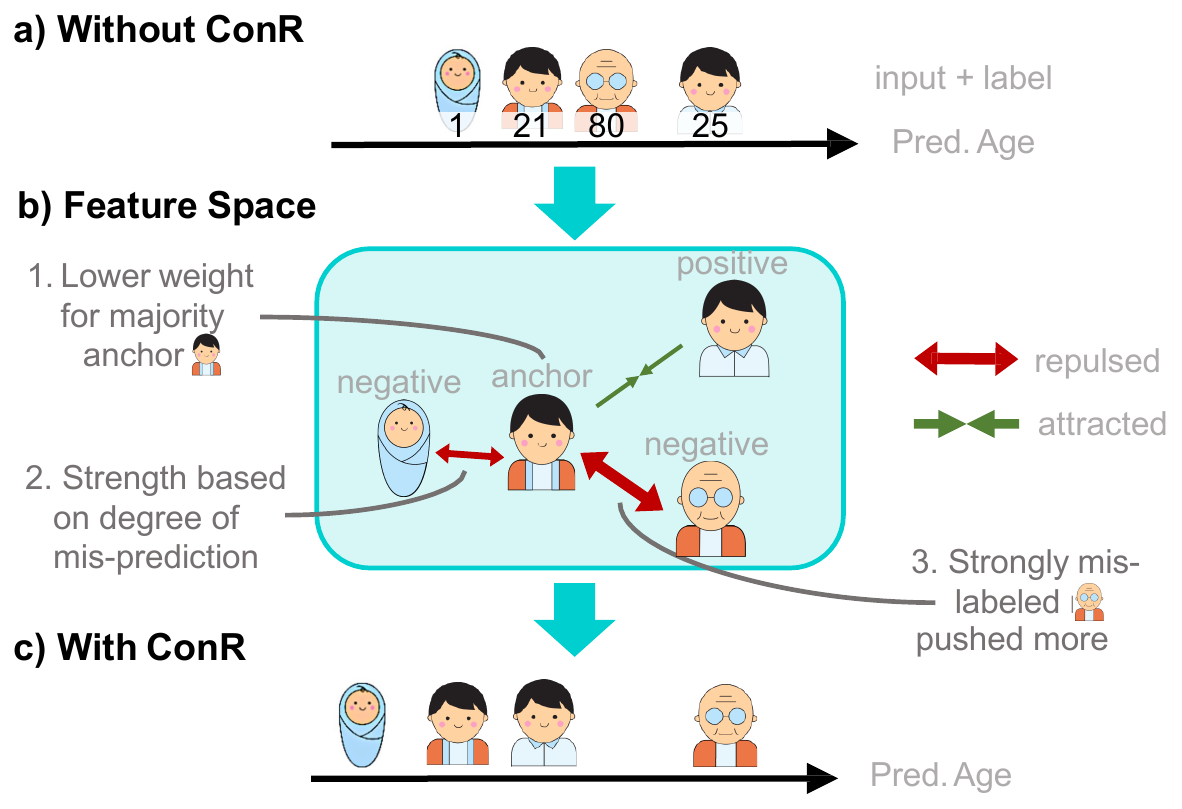}
    \caption{
    Key insights of \Pn{}.
    a) Without \Pn{}, it is common to have minority examples mixed with majority examples.
    b) \Pn{} selects the sample with confusion around it as an anchor and adjusts the feature space with relative contrastive learning.
    c) Reduced prediction error. 
    }
\vspace{-3mm}
    \label{fig: mot}
\end{wrapfigure}
We propose a method to enforce this correspondence: \Pn{} is a novel \textbf{Con}trastive \textbf{R}egularizer which is based on infoNCE loss~\citep{oord2018representation} but adapted for multi-positive pairs. While similar extensions are performed for classification tasks~\citep{supcon}, \Pn{} addresses continuous label space and penalizes minority sample features from collapsing into the majority ones.
Contrastive approaches for imbalanced classification are mainly based on 
predefined anchors, decision boundary-based negative pair selection and imposing discriminative feature space~\citep{cui2021parametric,wang2021contrastive,li2022targeted}. These are not feasibly extendible to continuous label spaces. \Pn{} introduces continuity to contrastive learning for regression tasks.
Fig~\ref{fig: mot} illustrates an intuitive example for \Pn{} from the task of age estimation. There are images of individuals of varying ages, including 1, 21, 25, and 80 years.  Age 1 and 80 are the minority examples, reflecting the limited number of images available for these age groups within the datasets. While 21 is a majority example, given the abundance of images around this age within the datasets.
Without using \Pn{},  similarities in the feature space are not aligned with the relationships in the label space. Thus, the minority samples' features collapse to the majority sample, leading to inaccurate predictions for minority ones that mimic the majority sample~(Fig~\ref{fig: mot}a).
\Pn{} regularizes the feature space by simultaneously encouraging locality via pulling together positive pairs and preserving global correlations by pushing negative pairs. The 21-year-old sample coexists within a region in the feature space alongside 1-year-old and 80-year-old samples. Thus, \Pn{} 1) considers the 21-year-old sample as an anchor,  and, 
2) pushes negative pairs for minority anchors harder than majority examples to provide better separation in feature space. Furthermore, \Pn{}
3) increases pushing power based on how heavily mislabelled an example is (Fig~\ref{fig: mot}b.1 to Fig~\ref{fig: mot}b.3). 
We demonstrate that \Pn{} effectively translates label relationships to the feature space, and boosts the regression performance on minority samples (Fig~\ref{fig: mot}c). Refer to~\ref{Empirical analysis of the motivation} for the empirical analysis on the motivation of \Pn{}.


\Pn{} implicitly models the local and global correlations in feature space by three main contributions to the contrastive objective:
{\textbf{1)~Dynamic anchor selection:}~Throughout the training, considering the learned proximity correspondences, \Pn{} selects samples with the most collapses on the feature manifold as anchors. 
\textbf{2)~Negative Pairs selection:}~Negative pairs are sampled to quantify the deviation they introduce to the correspondence of similarities in feature and label space, thereby compensating for under-represented samples.
\textbf{3)~Relative pushing:}~Negative pairs are pushed away proportional to their label similarity and the density of the anchor's label. 

\Pn{} is orthogonal to other imbalanced learning techniques and performs seamlessly on high-dimensional label spaces.
 Our comprehensive experiments on large-scale DIR benchmarks for facial age, depth and gaze estimation show that \Pn{} strikes a balance between efficiency and performance, especially on depth estimation, which has a complicated and high-dimensional label space.
\section{Related work}
\label{related work}
\textbf{Imbalanced classification.} The main focus of existing imbalanced approaches is on the classification task, while imbalanced regression is under-explored. The imbalanced classification methods are either data-driven or model-driven. Resampling~\citep{chawla2002smote,chu2020feature,byrd2019effect,han2005borderline,jiang2021improving} is a data-driven technique that balances the input data by either over-sampling~\citep{chawla2002smote,byrd2019effect} the minority classes or under-sampling the majority ones~\citep{han2005borderline}. Model-Aware K-center(MAK)~\citep{jiang2021improving}  over-samples tail classes using an external sampling pool. Another branch of data-driven approaches is augmentation-based methods. Mixup-based approaches linearly interpolate data either in input space~\citep{zhang2017mixup} or feature space~\citep{verma2019manifold} for vicinal risk minimization. RISDA~\citep{chen2022imagine} shares statistics between classes considering a confusion-based knowledge graph and implicitly augments minority classes. Model-driven methods such as focal loss~\citep{8237586} and logit-adjustment ~\citep{NEURIPS2020_5ca359ab} are cost-sensitive approaches that regulate the loss functions regarding the class labels.  To encourage learning an unbiased feature space,  several training strategies including two-stage training~\citep{kang2019decoupling}, and transfer learning~\citep{yin2019feature} are employed.
As discussed further, though there has been much success in this area, there are many issues in converting imbalanced classification approaches to regression.

\textbf{Imbalanced regression.}
Unlike classification tasks with the objective of learning discriminative representation, effective imbalanced learning in continuous label space is in lieu of modelling the label relationships in the feature space~\citep{yang2021delving,ranksim}. Therefore, imbalanced classification approaches do not feasibly extend to the continuous label space. 
 DenseLoss~\citep{steininger2021density} and LDS~\citep{yang2021delving} encourage local similarities by applying kernel smoothing in label space. Feature distribution smoothing (FDS)~\citep{yang2021delving} extends the idea of kernel smoothing to the feature space. Ranksim~\citep{ranksim} proposes to exploit both local and global dependencies by encouraging a correspondence of similarity order between labels and features. Balanced MSE~\citep{ren2022balanced} prevents Mean Squared Error (MSE) from  carrying imbalance to the prediction phase by restoring a balanced prediction distribution.
 The empirical observations in~\citep{yang2021delving} and highlighted in VIR~\citep{wang2023variational}, demonstrate that using empirical label distribution does not accurately reflect the real label density in regression tasks, unlike classification tasks. Thus, traditional re-weighting techniques in regression tasks face limitations. Consequently, LDS~\citep{yang2021delving} and the concurrent work, VIR propose to estimate effective label distribution. Despite their valuable success, two main issues arise, specifically for complicated label spaces. These approaches rely on binning the label space to share local statistics. However, while effective in capturing local label relationships, they disregard global correspondences. Furthermore, in complex and high-dimensional label spaces, achieving effective binning and statistics sharing demands in-depth domain knowledge. Even with this knowledge, the task becomes intricate and may not easily extend to different label spaces.
 
\textbf{Contrastive learning.}
Contrastive Learning approaches are pairwise representation learning techniques that push away semantically divergent samples and pull together similar ones~\citep{moco,pmlr-v119-chen20j,khosla2020supervised,kang2021exploring}. Momentum Contrast (Moco)~\citep{moco} is an unsupervised contrastive learning approach that provides a large set of negative samples via introducing a dynamic queue and a moving-averaged encoder; while SupCon~\citep{supcon} incorporates label information to the contrastive learning. \citet{kang2021exploring} argue that in case of imbalanced data, SupCon is subject to learning a feature space biased to majority samples and proposed k-positive contrastive loss (KCL) to choose the same number of positive samples for each instance to alleviate this bias. 
Contrastive long-tailed classification methods train parametric learnable class centres~\citep{cui2021parametric,wang2021contrastive} or encourage learning a regular simplex~\citep{li2022targeted}. However, these approaches cannot be used for regression tasks where handling label-wise prototypes is potentially complex. 
Contrastive Regression (CR)~\citep{wang2022contrastive} adds a contrastive loss to Mean Squared Error (MAE) to improve domain adaptation for the gaze estimation task. CR assumes there is a correspondence between similarities in label space and the feature space. Regardless, when learning from the imbalanced data, the correspondence can't be assumed.  Instead of the assumption of correspondence, \Pn\ translates the relative similarities from the label space to the feature space. \citet{barbano2023contrastive} proposed a relative pushing for the task of brain age prediction using MRI scans. This relative pushing doesn't consider imbalanced distribution, and the weights are learnable kernels that can introduce complexities to complex label spaces.
Rank-N-Contrast~\citep{zha2023rank} ranks samples and contrasts them against each other regarding their relative rankings to impose continuous representations for regression tasks. Like RankSim~\cite{ranksim}, Rank-N-Contrast is designed based on the order assumption and cannot be used in all regression tasks.


\section{Method}
\label{method}

\subsection{Problem definition}
Consider a training dataset 
consisting $N$ examples, which we denote as $\{(x_i,y_i)\}^N_{i=0}$, where $x_i \in 
 \mathbb{R}^{d}$ is an example input, and $y_i \in \mathbb{R}^{d'}$ is its corresponding label. $d$ and $d'$ are the dimensions of the input and label, respectively. 
We additionally enforce that the distribution of the labels $\mathcal{D}_y$ deviates significantly from a uniform distribution. Given a model that consists of a feature encoder $\mathcal{E}(\cdot)$, and a regression function $\mathcal{R}(\cdot)$, the objective is to train a regression model $\mathcal{R}(\mathcal{E(\cdot)})$, such that the model output $\hat{y}_i = \mathcal{R}(\mathcal{E}(x_i))$ is similar to the true label $y_i$.

 \subsection{Imbalanced contrastive regression}
In regression tasks, inter-label relationships unveil meaningful associations in feature space. 
However, learning from imbalanced data harms this correspondence since the features learned for minority samples share statistics with the majority samples, despite dissimilar labels~\citep{yang2021delving}.  By incorporating label and prediction relationships into contrastive learning, \Pn{} enforces appropriate feature-level similarities for modelling a balanced feature space. 
\begin{figure}[t]
    
  
  \includegraphics[width=\textwidth]{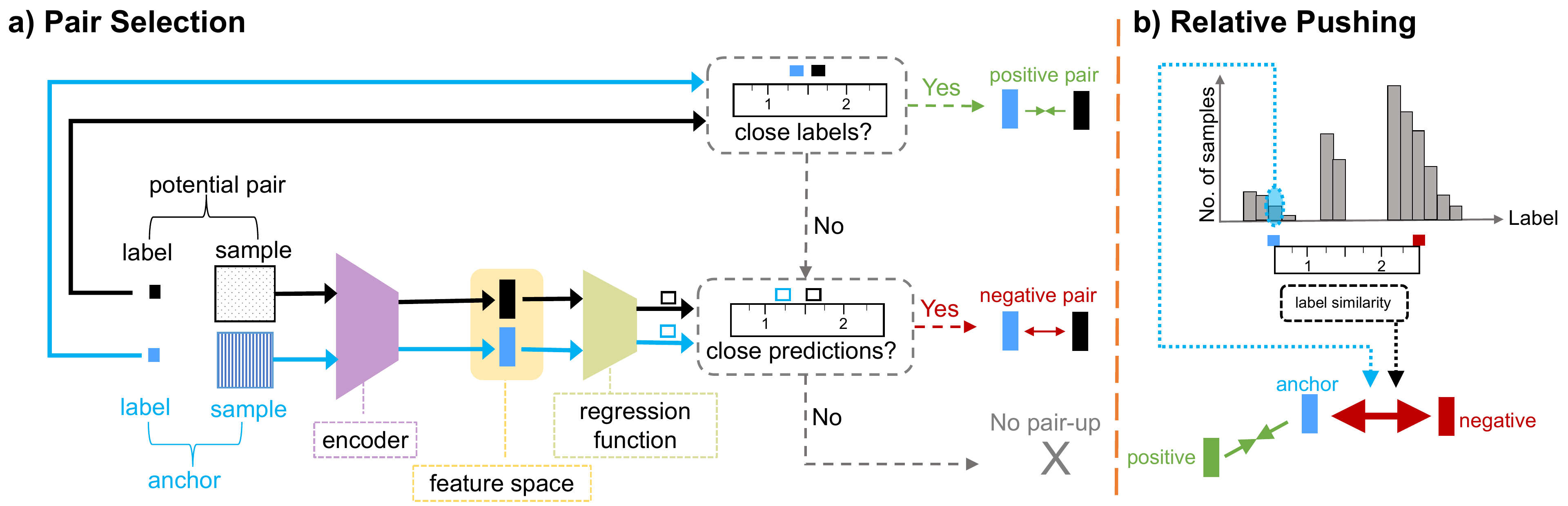}
  \vspace{-15pt}
  \caption{The framework of \Pn{} is to translate label similarities to the feature space. 
  a) Per each augmented sample, \Pn{} selects positive and negative pairs with regard to the label similarities and prediction similarities. b) \Pn{} pulls positive pairs together while pushing away negative pairs regarding their label similarities and label distribution for the anchor. In this way, the minority anchor pushes negative samples harder. The pushing weight is inversely relative to the label similarities.
  }
  \label{Figure_arch}
\end{figure}

\Pn{} is a continuous variation of infoNCE. Initially, for creating diversity in examples, we perform problem-specific augmentations on each input $x_i$ to produce two augmented samples. We define the set of augmented examples from the input to be $\{(x^a_j,y_j)\}^{2N}_{j=0}$, 
where $x^a_j$ is an augmented input. 

As illustrated in Fig.~\ref{Figure_arch}, \Pn{} is a collaboration of pair selection and relative pushing.
For each augmented sample, ConR first selects pairs. Next, in the case of at least one negative pairing, the sample is considered an anchor and contributes to the regularizing process of \Pn{} by pulling together positive pairs and relatively repelling negative pairs. 
\subsubsection{Pair selection}
\vspace{-2mm}
Given a pair of examples $(x^a_i, y_i)$ and $(x^a_j, y_j)$ from the augmented inputs (\textit{labels} and \textit{samples}, Fig.~\ref{Figure_arch}-a), 
each example is passed to the feature encoder $\mathcal{E}(\cdot)$ to produce feature vectors $z_i$ and $z_j$, and then to the regression function $\mathcal{R}(\cdot)$ for predictions $\hat{y}_i$ and $\hat{y}_j$ (\textit{sample} through \textit{regression function}, Fig.~\ref{Figure_arch}-a).
The values of the predicted and actual labels are used to determine if the examples should be a positive pair, a negative pair, or unpaired (righthand side, Fig.~\ref{Figure_arch}-a). 

To measure similarity between labels (or predictions) of augmented examples, \Pn{} defines a similarity threshold $\omega$. 
Given a similarity function $Sim(\cdot, \cdot) \in \mathbb{R} $ (e.g., inverse square error), we define two labels (or two predictions) $y_i$ and $y_j$ as similar if $Sim(y_i, y_j) \geq \omega$.  We denote this as $y_i \simeq y_j$. 
Iff two examples have similar labels $y_i \simeq y_j$, then they are treated as a positive pair. The examples have dissimilar labels, but similar predictions $\hat{y}_i \simeq \hat{y}_j$, then they are treated as a negative pair. Otherwise, examples are unpaired.
\paragraph{Anchor selection.}
For each example $j$, $(x^a_j, y_j)$
We denote the sets of positive and negative pairs for this example, and their feature vectors as $K^+_j = \{(z_p)\}^{N^+_j}_p$ and $K^-_j = \{(z_q)\}^{N^-_j}_q$,
respectively, where $N^+_j$ is the number of positive examples and $N^-_j$ is the number of negative samples for example $j$. 
 If $N^-_j > 0$, $(x^a_j, y_j)$ is selected as an anchor and contributes to the regularization process of \Pn{}. 
\subsubsection{Contrastive regularizer} 
For each example $j$, \Pn\ introduces a loss function ${\mathcal{L}_{ConR}}_j$. 
If example $j$ is not selected as an anchor, ${\mathcal{L}_{ConR}}_j =0$.  Otherwise, ${\mathcal{L}_{ConR}}_j$ pulls together positive pairs while pushing away negative pairs proportional to the similarity of their labels. As shown in Fig.~\ref{Figure_arch}-b, \Pn\ pushes away negative samples with less label similarity to $y_j$ harder than negative samples with labels closer to the anchor $y_j$:
\begin{equation}
     {\mathcal{L}_{ConR}}_j = -\log \frac{1}{N^+_j}
         \sum_{z_i \in K^+_j} \frac{
             \exp(z_j\cdot z_i/\tau)
         }{\sum_{z_p \in K^+_j} 
                 \exp(z_j\cdot z_p/\tau) 
             + \sum_{z_q \in K^-_j}
                \mathcal{S}_{j,q} \exp(z_j\cdot z_q/\tau)
        }
        \label{eq: conr}
\end{equation}
where $\tau$ is a temperature hyperparameter and $\mathcal{S}_{j,n}$ is a pushing weight for each negative pair:
\begin{equation}
    \label{eq: sim}
    \mathcal{S}_{j,q} = f_{\mathcal{S}}(\eta_j, Sim(y_j,y_q)), 
\end{equation}
and $y_q$ is the label of $x_q$ where $z_q = \mathcal{E}(x_q)$. $\eta_j$ is a pushing power for each sample $(x^a_j,y_j)$ that depends on label distribution $\mathcal{D}_y$ to boost the pushing power for minority samples~(Fig.~\ref{Figure_arch}-b): $\eta_j \propto w_j$, where $w_j$ is a density-based weight for input $j$ derived from the empirical label distribution (e.g., inverse frequency). The function $f_\mathcal{S}$ computes the $\mathcal{S}_{j,q} $ to be proportionate to $\eta_j$ and inversely related to $Sim(y_j,y_q)$. Please refer to Appendix~\ref{implementation details} for the definition of $f_\mathcal{S}$.

Finally, $\mathcal{L}_{ConR}$ is the \Pn's regularizer value for the augmented example set:
\begin{equation}
   \mathcal{L}_{ConR} = \frac{1}{2N}\sum^{2N}_{j=0}{\mathcal{L}_{ConR}}_j.
   \label{eq:conr-all}
\end{equation}
To prevent the representations of minority labels from collapsing to the majority ones in deep imbalanced regression, $L_{sum}$ is optimized. $L_{sum}$ is weighed sum of  $\mathcal{L}_{ConR}$ and a regression loss $\mathcal{L}_\mathcal{R}$ (e.g., mean absolute error) as below:
\begin{equation}
    \label{eq: sum}
    L_{sum} = \alpha \mathcal{L}_\mathcal{R} + \beta \mathcal{L}_{ConR}
    \vspace{-10pt}
\end{equation}
\subsection{Theoretical insights}
We theoretically justify the effectiveness of \Pn{} in Appendix \ref{appendix:theoretical}. We derive the upper bound $\mathcal{L}_{ConR} + \epsilon$ on the probability of incorrect labelling of minority samples to be:
\begin{align}
    \frac{1}{4N^2}\sum^{2N}_{j=0, x_j \in A} \sum^{ K^-_j}_{q=0}\log\mathcal{S}_{j,q}
                 p(\hat Y_j|x_q)  \le \mathcal{L}_{ConR} + \epsilon, \;\; \epsilon \overset{N\rightarrow \infty}{\rightarrow}  0   
\end{align}
where $A$ is the set of anchors and $x_q$ is a negative sample. $p(\hat Y_j|x_q)$ is the likelihood of sample $x_q$ with an incorrect prediction $y_q \in \hat Y_j = (\hat y_j - \omega, \hat y_j + \omega)$. $\hat y_j$ a prediction that is mistakenly similar to its negative pair $\hat y_q$. We refer to $p(\hat Y_j|x_q)$ as the probability of collapse for $x_q$.
The Left-Hand Side (LHS) contains the probability of all collapses for all negative samples, which is bounded above by  $\mathcal{L}_{ConR}$.
Minimizing the loss, consequently minimizes the LHS, which causes either 1) the number of anchors decreases or 2) the degree of collapses is reduced. 
Each $p(\hat Y_j|x_q)$ is weighted by $\mathcal{S}_{j,q}$, which penalizes incorrect predictions proportional to severity. 
Thus, optimizing ConR decreases the probability of mislabelling proportional to the degree, and emphasizes minority samples.
\color{black}
\section{Experiments}
\vspace{-2mm}
\label{experiment}

\subsection{Main results}
\vspace{-2mm}
\paragraph{Datasets and baselines.}
We use three datasets curated by \citet{yang2021delving} for the deep imbalanced regression problem: AgeDB-DIR is a facial age estimation benchmark, created based on AgeDB~\citep{moschoglou2017agedb}. 
IMDB-WIKI-DIR is an age estimation dataset originated from IMDB-WIKI~\citep{rothe2018deep}. 
NYUD2-DIR is created based on NYU Depth Dataset V2~\citep{silberman2012indoor} to predict the depth maps from RGB indoor scenes. Moreover, we create MPIIGaze-DIR based on MPIIGaze, which is an appearance-based gaze estimation benchmark. Refer to Appendix \ref{dataset details} and \ref{implementation details} for more dataset and implementation details. Please refer to Appendix \ref{appendix: baselines} for baseline details.

\paragraph{Evaluation process and metrics.}

Following the standard procedure for imbalanced learning~\citep{yang2021delving,ranksim}, we report the results for four shots: \textit{All}, \textit{few}, \textit{median} and \textit{many}. The whole test data is denoted by \textit{All}. Based on the number of samples in the training dataset, \textit{few}, \textit{median} and \textit{many} that have less than 20, between 20 and 100, and more than 100 training samples per label, respectively.

For AgeDB-DIR and IMDB-WIKI-DIR, the metrics are Mean-Absolute-Error~(MAE), and Geometric Mean~(GM). For NYUD2-DIR we use Root Mean Squared Error (RMSE) and Threshold accuracy ($\delta_1$) as metrics as in ~\citep{yang2021delving,ren2022balanced}. Threshold accuracy $\delta_i$ is the percentage of $d_i$ that satisfies $max(\frac{d_1}{g_1},\frac{g_1}{d_1})  < {1.25}$, where for each pixel, $g_1$ is the ground truth depth value and $d_1$ is the predicted depth value. For MPIIGaze-DIR, we use Mean Angle Error (degrees).
To calculate relative improvement of \Pn{}, we compare the performance of each combination of methods including \Pn{}, against the same combination without \Pn{}.
Each "Ours \textit{vs.} ..." entry shows the average of these improvements~(e.g. "Ours \textit{vs.} LDS" is the average of the improvements of adding \Pn{} to each combination of baselines that has LDS).

\paragraph{Main results for age estimation.}
Table~\ref{table: agedb} and Table~\ref{table: imdb} show the results on AgeDB-DIR and IMDB-WIKI-DIR benchmarks, respectively. We compare the performance of DIR methods: FDS, LDS and RankSim with their regularized version by \Pn{}. All results are the averages of 5 random runs.
We observe that the performances of all the baselines are considerably boosted when they are regularized with \Pn{}. In addition, \Pn{} results in the best performance across both metrics and all shots for the AgeDB-DIR benchmark with leading MAE results of 6.81 and 9.21 on the overall test set and \textit{few}-shot region, respectively. For IMDB-WIKI-DIR, \Pn{} achieves the highest performance on 7 out of 8 shot/metric combinations with the best MAE results of 7.29 and 21.32 on the overall test set and \textit{few}-shot region, respectively.

\begin{table}[h]

    \caption{\textbf{Main results for AgeDB-DIR benchmark}. Results are reported for the whole test data (\textit{all}) and three other shots: \textit{many}, \textit{median}, and \textit{few}. Each section compares a baseline with its regularized version of \Pn. At the bottom of the table, the average improvements with respect to corresponding baselines without \Pn{} are reported in \textcolor{green}{Green}. The best result of either a baseline or its regularized version by \Pn{} is in \textbf{bold} (in column) and \textcolor{red}{\textbf{Red}} (across column). Baselines:~\citep{ranksim}.}
    \resizebox{\textwidth}{!}{%
    \centering
    \label{table: agedb}
    \begin{tabular}{l?cccc@{\hspace{40pt}}cccc}
    
        \specialrule{2\heavyrulewidth}{\abovetopsep}{\belowbottomsep}
        \textbf{Metrics}&\multicolumn{4}{c}{\textbf{MAE}$\downarrow$}&\multicolumn{4}{c}{\textbf{GM}$\downarrow$}\\
        \specialrule{1.5\heavyrulewidth}{\abovetopsep}{\belowbottomsep}
        \textbf{Methods/Shots} & \textbf{All} & \textbf{Many} & \textbf{Median} & \textbf{Few} & \textbf{All} & \textbf{Many} & \textbf{Median} & \textbf{Few}\\

        \specialrule{1.5\heavyrulewidth}{\abovetopsep}{\belowbottomsep}
        \Pn-only~\textbf{(Ours)} & 7.20&6.50&8.04&9.73 &4.59&3.94&4.83&6.39 \color{black}\\
        
        \hline
        
         LDS &  7.42 & 6.83& 8.21 & 10.79&4.85&4.39&5.80&7.03\\
         LDS + \Pn~\textbf{(Ours)} &\textbf{7.16}&\textbf{ 6.61} &\textbf{7.97}&\textbf{9.62} &\textbf{4.51}&\textbf{4.21}&\textbf{4.92}&\textbf{\textcolor{red}{5.87}}\\
        \hline
          FDS& 7.55&6.99&8.40& 10.48 &4.82&4.49&5.47&6.58\\
          FDS +  \Pn~\textbf{(Ours)} &\textbf{7.08}&\textbf{6.46}&\textbf{7.89}&\textbf{9.80}&\textbf{4.31}&\textbf{4.01}&\textbf{5.25}&\textbf{6.92}\\
         \hline
          RankSim & 6.91&6.34&7.79& 9.89&4.28&3.92&4.88&6.89\\
         RankSim + \Pn~\textbf{(Ours)}&\textbf{6.84}&\textbf{6.31}&\textbf{7.65}&\textbf{9.72}&\textbf{4.21}&\textbf{3.95}&\textbf{4.75}&\textbf{6.28}\\
        \hline
         LDS + FDS &7.37 &\textbf{6.82} &8.25&10.16 &4.72& \textbf{4.33} &5.50 &6.98\\       
         LDS + FDS + \Pn~\textbf{(Ours)} &\textbf{7.21}&6.88&\textbf{7.63}&\textbf{9.59}&\textbf{4.63}&4.45&\textbf{5.18}&\textbf{5.91}\\
        \hline
        
         LDS + RankSim&6.99& 6.38 &7.88&10.23 &\textbf{4.40}&3.97&5.30&6.90\\
          LDS + RankSim + \Pn~\textbf{(Ours)}&\textbf{6.88}&\textbf{6.35}&\textbf{7.69}&\textbf{9.99}&4.43&\textbf{3.87}&\textbf{4.70}&\textbf{6.51}\\
         \hline
        FDS + RankSim&7.02 &6.49& 7.84 &9.68 &4.53 &4.13 &5.37 &6.89\\
        FDS + RankSim+ \Pn~\textbf{(Ours)}&\textbf{6.97}&\textbf{6.33}&\textbf{7.71}&\textbf{9.43}&\textbf{\textcolor{red}{4.19}}&\textbf{3.92}&\textbf{\textcolor{red}{4.67}}&\textbf{6.14}\\
       \hline
        LDS + FDS + RankSim &7.03 &6.54 &7.68 &9.92 &4.45 &4.07 &5.23 &6.35\\

        LDS + FDS + RankSim + \Pn~\textbf{(Ours)}&\textbf{\textcolor{red}{6.81}}&\textbf{\textcolor{red}{6.32}}&\textbf{\textcolor{red}{7.45}}&\textbf{\textcolor{red}{9.21}}&\textbf{4.39}&\textbf{\textcolor{red}{3.81}}&\textbf{5.01}&\textbf{6.02}\\
        
        \hline\hline
        \textbf{Ours} \textit{vs.} LDS &\textcolor{green}{2.58\%}&	\textcolor{green}{1.51\%}&	\textcolor{green}{3.97\%}&	\textcolor{green}{6.55\%}&	\textcolor{green}{2.39\%}&	\textcolor{green}{2.43\%}&	\textcolor{green}{9.15\%}&	\textcolor{green}{10.78\%}\\
        \textbf{Ours} \textit{vs.} FDS &\textcolor{green}{3.07\%}	&\textcolor{green}{3.14\%}&	\textcolor{green}{4.57\%}&	\textcolor{green}{5.47\%}&	\textcolor{green}{5.34\%}&\textcolor{green}{4.85\%}&	\textcolor{green}{6.78	\%}&\textcolor{green}{6.56\%}\\
        \textbf{Ours} \textit{vs.} RankSim &\textcolor{green}{1.62\%}	&\textcolor{green}{1.70\%}&	\textcolor{green}{2.24\%}	&\textcolor{green}{3.49\%}	&\textcolor{green}{2.45\%}	&\textcolor{green}{3.31\%}	&\textcolor{green}{8.14\%}	&\textcolor{green}{7.79\%}\\
        \specialrule{2\heavyrulewidth}{\abovetopsep}{\belowbottomsep}
        
        
    \end{tabular}
    
    }
\end{table}
\begin{table}[htb!]    
    \caption{\textbf{Main results on IMDB-WIKI-DIR benchmark.}}
    \resizebox{\textwidth}{!}{%
    \centering
    \label{table: imdb}
    \begin{tabular}{l?
    cccc@{\hspace{40pt}}cccc}
    
        \specialrule{2\heavyrulewidth}{\abovetopsep}{\belowbottomsep}
        \textbf{Metrics}&\multicolumn{4}{c}{\textbf{MAE}$\downarrow$}&\multicolumn{4}{c}{\textbf{GM}$\downarrow$}\\
        \specialrule{1.5\heavyrulewidth}{\abovetopsep}{\belowbottomsep}
        \textbf{Methods/Shots} & \textbf{All} & \textbf{Many} & \textbf{Median} & \textbf{Few} & \textbf{All} & \textbf{Many} & \textbf{Median} & \textbf{Few}\\
        \specialrule{1.5\heavyrulewidth}{\abovetopsep}{\belowbottomsep}
         \Pn-only~\textbf{(Ours)}&7.33	&6.75	&11.99	&22.22&4.02&3.79&6.98&12.95 \color{black}\\
        \hline
        
         LDS &7.83&7.31&12.43&22.51&4.42 &4.19& 7.00 &13.94\\
         LDS + \Pn~\textbf{(Ours)} &\textbf{7.43}&\textbf{6.84}&\textbf{12.38}& \textbf{21.98}&\textbf{4.06}&\textbf{3.94}&\textbf{6.83}&\textbf{12.89}\\
        \hline
         FDS &7.83&7.23&12.60&22.37 &4.42 &4.20 &6.93& 13.48\\
          FDS + \Pn~\textbf{(Ours)} &\textbf{\textcolor{red}{7.29}}&\textbf{6.90}&\textbf{12.01}&\textbf{21.72}&\textbf{4.02}&\textbf{3.83}&\textbf{6.71}&\textbf{12.59}  \\
         \hline
          RankSim &7.42&6.84&12.12& 22.13 &4.10 &3.87 &6.74 &12.78\\
         RankSim + \Pn~\textbf{(Ours)}&\textbf{7.33}	&\textbf{\textcolor{red}{6.69}}&	\textbf{11.87}	&\textbf{21.53}	&\textbf{\textcolor{red}{3.99}}&	\textbf{\textcolor{red}{3.81}}&	\textbf{6.66}	&\textbf{12.62}\\
        \hline
         LDS + FDS &7.78 &7.20& 12.61 &22.19& 4.37 &4.12 &7.39& \textbf{12.61}\\     
         LDS + FDS + \Pn~\textbf{(Ours)} &\textbf{7.37}&\textbf{7.00}&\textbf{12.31}&\textbf{21.81}&\textbf{4.07}&\textbf{3.96}&\textbf{6.88}& 12.86 \\
        \hline
        
         LDS + RankSim &7.57&7.00&12.16&22.44&4.23 &4.00 &6.81 &13.23  \\
          LDS + RankSim + \Pn~\textbf{(Ours)}&\textbf{7.48}&	\textbf{6.79}&	\textbf{12.03}	&\textbf{22.31}	&\textbf{4.04}&	\textbf{3.86}&	\textbf{6.77}	&\textbf{12.80}\\
         \hline
        FDS + RankSim &7.50 &6.93 &12.09& 21.68& 4.19 &3.97 &6.65 &13.28  \\
        FDS + RankSim+ \Pn~\textbf{(Ours)}&\textbf{7.44}	&\textbf{6.78}&	\textbf{\textcolor{red}{11.79}}	&\textbf{\textcolor{red}{21.32}}	&\textbf{4.05}&	\textbf{3.88}	&\textbf{\textcolor{red}{6.53}}	&\textbf{12.67}\\
       \hline
        LDS + FDS + RankSim &7.69& 7.13 &12.30& 21.43& 4.34 &4.13 &6.72 &\textbf{\textcolor{red}{12.48}}  \\

        LDS + FDS + RankSim + \Pn~\textbf{(Ours)}&\textbf{7.46}&	\textbf{6.98}	&\textbf{12.25}&	\textbf{21.39}&	\textbf{4.19}&	\textbf{4.01}	&\textbf{6.75}	&12.54\\
        
        \hline\hline
        
        \textbf{Ours} \textit{vs.} LDS &\textcolor{green}{3.67\%}&	\textcolor{green}{3.54\%}&	\textcolor{green}{1.07\%}&	\textcolor{green}{1.21\%}&	\textcolor{green}{5.75\%}&	\textcolor{green}{4.07\%}&	\textcolor{green}{2.37\%}&	\textcolor{green}{2.08\%}\\
        \textbf{Ours} \textit{vs.} FDS &\textcolor{green}{4.00\%}	&\textcolor{green}{2.91\%}&	\textcolor{green}{2.49\%}&	\textcolor{green}{1.62\%}&	\textcolor{green}{5.68\%}&\textcolor{green}{4.47\%}&	\textcolor{green}{2.86	\%}&\textcolor{green}{2.18\%}\\
        \textbf{Ours} \textit{vs.} RankSim &\textcolor{green}{1.55\%}	&\textcolor{green}{2.37\%}&	\textcolor{green}{1.51\%}	&\textcolor{green}{1.29\%}	&\textcolor{green}{3.50\%}	&\textcolor{green}{2.56\%}	&\textcolor{green}{0.79\%}	&\textcolor{green}{2.16\%}\\
        \specialrule{2\heavyrulewidth}{\abovetopsep}{\belowbottomsep}

        
    \end{tabular}
    
    }
\end{table}

\paragraph{Main results for depth estimation.} 
To assess the effectiveness of \Pn{} in more challenging settings where the label space is high-dimensional with non-linear inter-label relationships, we compare its performance with baselines: LDS, FDS and Balanced MSE on NYUD2-DIR depth estimation benchmark. 
For this task, the label similarity is measured by the difference between the average depth value of two samples. As shown in table~\ref{table: NYU}, 
\Pn{} alleviates the bias of the baseline toward the majority samples. 
\Pn{} significantly outperforms LDS, FDS and Balanced MSE across all shots and both metrics with leading RMSE results of 1.265 on the overall test set and 1.667 on the \textit{few}-shot region. Notably, RankSim cannot be used on the depth estimation task; however, \Pn{} can smoothly be extended to high-dimensional label space with high performance. The reason is that order relationships are not semantically meaningful for all regression tasks with high-dimensional label space, such as depth estimation.
\begin{table}[htb!]
\caption{\textbf{Main results on NYUD2-DIR benchmark.} }
\vspace{-5pt}
    \resizebox{\textwidth}{!}{%
    \centering
    
    \label{table: NYU}
    \begin{tabular}{l?cccc@{\hspace{30pt}}cccc}
        \specialrule{2\heavyrulewidth}{\abovetopsep}{\belowbottomsep}
        \textbf{Metrics}&\multicolumn{4}{c}{\textbf{RMSE}$\downarrow$}&\multicolumn{4}{c}{\textbf{$\delta_1$}$\uparrow$}\\
        \specialrule{1.5\heavyrulewidth}{\abovetopsep}{\belowbottomsep}
        \textbf{Methods/Shots} & \textbf{All} & \textbf{Many} & \textbf{Median} & \textbf{Few} & \textbf{All} & \textbf{Many} & \textbf{Median} & \textbf{Few}\\
        \specialrule{1.5\heavyrulewidth}{\abovetopsep}{\belowbottomsep}
        \Pn-only~\textbf{(Ours)}&1.304&0.682&0.889&1.885&0.675&0.699&0.753&0.648 \color{blue}\\
        \hline
         FDS & 1.442 &0.615 &0.940 &2.059 &0.681 &0.760 &0.695 &0.596\\
         FDS + \Pn{}~\textbf{(Ours)}&\textbf{1.299}	&\textbf{0.613}	&\textbf{0.836}&	\textbf{1.825}&	\textbf{0.696}&	\textbf{\textcolor{red}{0.800}}&	\textbf{0.819}&	\textbf{0.701}\\
        \hline
         LDS &1.387 &\textbf{0.671} &0.913 &1.954 &0.672 &\textbf{0.701}& 0.706 &0.630\\
         LDS + \Pn ~\textbf{(Ours)}&\textbf{1.323}&0.786&\textbf{0.823}&\textbf{1.852}&\textbf{0.700}&0.632&\textbf{0.827}&\textbf{0.702}\\
        \hline
         RankSim &-&-&-&-&-&-&-&-\\
        \hline
        Balanced MSE~(BNI) &  1.283 &0.787 &0.870 &1.736 &0.694 &0.622 &0.806 &\textbf{0.723}\\
        
         Balanced MSE~(BNI) + \Pn{}~\textbf{(Ours)}&\textbf{\textcolor{red}{1.265}}&\textbf{0.772}&\textbf{\textcolor{red}{0.809}}&\textbf{1.689\color{black}}&\textbf{\textcolor{red}{0.705}}&\textbf{0.631}&\textbf{\textcolor{red}{0.832}}&0.698\\
        \hline
         Balanced MSE~(BNI) + LDS &1.319	&0.810	&0.920&	1.820&	0.681	&0.601	&0.695	&0.648\\
         Balanced MSE~(BNI) + LDS + \Pn{}~\textbf{(Ours)}&\textbf{1.271}&\textbf{0.723}&\textbf{0.893}&\textbf{\textcolor{red}{1.667}}&\textbf{0.699}&\textbf{0.652}&\textbf{0.761}&\textbf{\textcolor{red}{0.752}}\\
        
        \hline\hline
        \textbf{Ours} \textit{vs.} LDS&\textcolor{green}{4.13\%}&\textcolor{green}{4.25\%}&\textcolor{green}{10.41\%}&\textcolor{green}{6.81\%}&\textcolor{green}{3.41\%}&\textcolor{green}{0.07\%}&\textcolor{green}{13.31\%}&\textcolor{green}{13.74\%}\\
        \textbf{Ours} \textit{vs.} FDS&\textcolor{green}{9.92\%}&\textcolor{green}{0.03\%}&\textcolor{green}{11.06\%}&\textcolor{green}{11.37\%}&\textcolor{green}{2.20\%}&\textcolor{green}{5.27\%}&\textcolor{green}{17.84\%}&\textcolor{green}{17.62\%}\\
        \textbf{Ours} \textit{vs.} Balanced MSE~(BNI)&\textcolor{green}{2.52\%}&\textcolor{green}{6.33\%}&\textcolor{green}{8.99\%}&\textcolor{green}{18.42\%}&\textcolor{green}{2.11\%}&\textcolor{green}{4.97\%}&\textcolor{green}{6.36\%}&\textcolor{green}{6.30\%}\\
        \specialrule{2\heavyrulewidth}{\abovetopsep}{\belowbottomsep}

    \end{tabular}  
    }
    \vspace{-5mm}
\end{table}

\begin{table}[htb!]   
\vspace{-5pt}
    \caption{\textbf{Main results of on MPIIGaze-DIR benchmark.}}
    \vspace{-2pt}
    \centering
    \resizebox{.7\textwidth}{!}{%
    \centering
    \label{table: gaze-all}
    \begin{tabular}{l?cccc}
    
        \specialrule{1.5\heavyrulewidth}{\abovetopsep}{\belowbottomsep}
        \textbf{Metric}&\multicolumn{4}{c}{\textbf{Mean angular error (degrees)}$\downarrow$}\\
        \specialrule{1.5\heavyrulewidth}{\abovetopsep}{\belowbottomsep}
        \textbf{Method/Shot} & \textbf{All} & \textbf{Many} & \textbf{Median} & \textbf{Few} \\
        
        \specialrule{1.5\heavyrulewidth}{\abovetopsep}{\belowbottomsep}
         \Pn{}-only~\textbf{(Ours)}&6.16&	5.73&	6.85&	6.17 \color{blue}\\
        \hline
        LDS &6.48&	5.93&	7.28&	6.83\\
        LDS + \Pn{}~\textbf{(Ours)}&\textbf{6.08}&	\textbf{5.76}&	\textbf{6.81}&	\textbf{5.98} \\
        \hline
        FDS &6.71&	6.01&	7.50&	6.64\\
        FDS + \Pn{}~\textbf{(Ours)}&\textbf{6.32}&	\textbf{5.69}&	\textbf{6.56}&	\textbf{5.65} \\
        \hline
        Balanced MSE (BNI) &5.73&	6.34&	6.41&	5.36\\
        Balanced MSE (BNI)+ \Pn{}~\textbf{(Ours)}&\textbf{\textcolor{red}{5.63}}&	\textbf{\textcolor{red}{6.02}}&	\textbf{\textcolor{red}{6.17}}&	\textbf{\textcolor{red}{5.21}}\\
        \hline\hline
        \textbf{Ours} vs. LDS & \textcolor{green}{6.17 \%}   & \textcolor{green}{2.87\%}   & \textcolor{green}{6.46\%}   & \textcolor{green}{12.45 \%}\\
        \textbf{Ours} vs. FDS & \textcolor{green}{5.81 \%}   & \textcolor{green}{5.32\%}   & \textcolor{green}{12.53\%}   & \textcolor{green}{14.91 \%}\\
        \textbf{Ours} vs. Balanced MSE & \textcolor{green}{1.75 \%}   & \textcolor{green}{5.05\%}   & \textcolor{green}{3.75\%}   & \textcolor{green}{2.80 \%}\\
        
        \specialrule{1.5\heavyrulewidth}{\abovetopsep}{\belowbottomsep}
        
    \end{tabular}
    }
\end{table}
\paragraph{Main results for gaze estimation.} Table~\ref{table: gaze-all} compares the performance of \Pn{} with three baseline methods: LDS, FDS and Balanced MSE on the MPIIGaze-DIR benchmark.  As shown in table~\ref{table: gaze-all}, 
\Pn{}  consistently improve the performance of all baselines for all shots and achieves the best Mean angular error of 5.63~(degrees) on the overall test set and 5.21~(degrees) on the \textit{few}-shot region.

\paragraph{Error reduction.} In Fig. ~\ref{fig: p-c}, we show the comparison between three strong baselines (LDS, FDS and Balanced MSE) and by adding \Pn{} for NYUD2-DIR benchmark. It shows a consistent and notable error reduction effect by adding our \Pn{} to the deep imbalanced learning. For more results and analysis on other datasets, please refer to Appendix ~\ref{Performance Analysis}.

\begin{figure}[t]
  \centering
  
  \begin{subfigure}{0.32\textwidth}
    \includegraphics[width=1\linewidth,height =32mm]{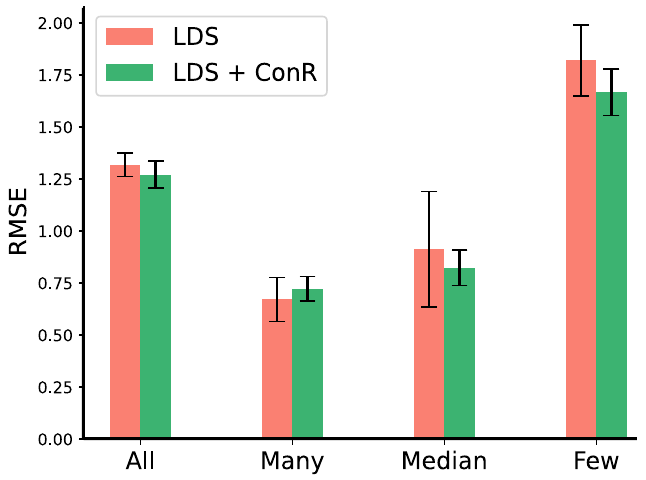}
  \end{subfigure}
  \hfill
  \begin{subfigure}{0.32\textwidth}   \includegraphics[width=1\linewidth,height =32mm]{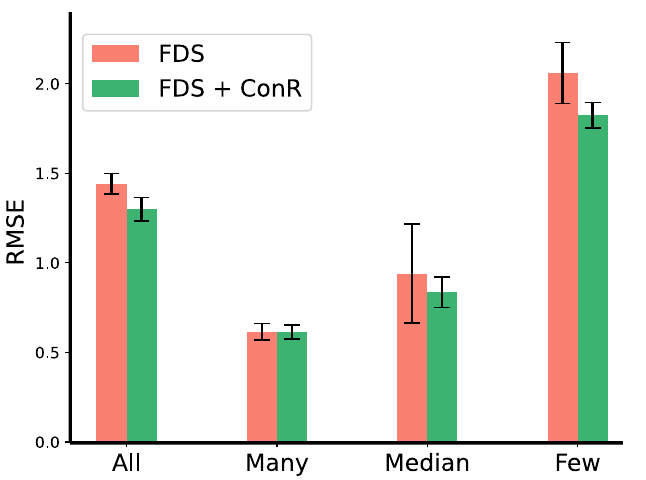}
  \end{subfigure}
  \hfill
  \begin{subfigure}{0.32\textwidth}  \includegraphics[width=1\linewidth,height =32mm]{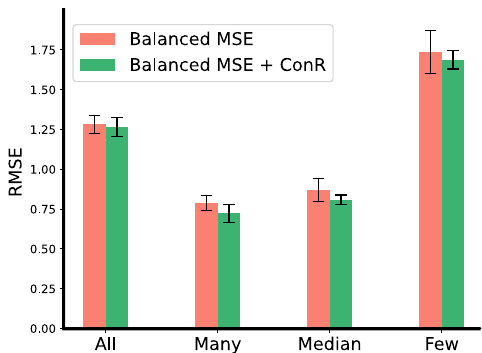}
  \end{subfigure}
  \vspace{-2mm}
  \caption{comparison on RMSE results by adding \Pn{} on top of the baselines for NYUD2-DIR benchmark.}
    \vspace{-5mm}
  \label{fig: p-c}
\end{figure}
\paragraph{Time consumption Analysis.}
Table~\ref{table: eff} provides the time consumption of \Pn{} in comparison to other baselines and VANILLA for the age estimation and depth estimation tasks. VANILLA is a regression model with no technique for imbalanced learning. The reported time consumptions are expressed in seconds for AgeDB-DIR and in minutes for NYUD2-DIR, representing the average forward pass and training time, and were measured using four NVIDIA GeForce GTX 1080 Ti GPUs. 
Table~\ref{table: eff}'s findings demonstrate that even with a high-dimensional label space, ConR's training time is considerably lower than FDS while remaining comparable to the time complexity of LDS, RankSim, and Balanced MSE. This indicates that ConR is an efficient alternative to FDS without compromising efficiency compared to other well-established methods. This result highlights ConR's ability to handle complex tasks without introducing significant computational overhead.
\begin{table}[h]    
    \vspace{-5pt}
    \caption{\textbf{Time consumption for AgeDB-DIR and NYU2D-DIR benchmarks.}}
    \vspace{-6pt}
    \centering
    \resizebox{.95\textwidth}{!}{%
    
    \label{table: eff}
    \begin{tabular}{l?cc:cc}
    
        \specialrule{1.5\heavyrulewidth}{\abovetopsep}{\belowbottomsep}

        \textbf{Benchmark}&\multicolumn{2}{c:}{\textbf{AgeDB-DIR}}&\multicolumn{2}{c}{\textbf{NYUD2-DIR}}\\
        \specialrule{1.5\heavyrulewidth}{\abovetopsep}{\belowbottomsep}
        \textbf{Method/Metric}&\textbf{Forward pass~(s)} & \textbf{Training time~(s)}&\textbf{Forward pass~(m)} & \textbf{Training time~(m)}\\
        
        
        \specialrule{1.5\heavyrulewidth}{\abovetopsep}{\belowbottomsep}
        VANILLA &12.2&31.5&7.5&28.3\\
        LDS &12.7&34.3&7.9&30.2\\
        FDS &38.4&60.5&10.8&66.3\\
        RankSim &16.8&38.8&-&-\\
        Balanced MSE &16.2&36.3&7.9&29.4\\
        \Pn{} (Ours)&15.7&33.4&8.1&30.4\\
        \specialrule{1.5\heavyrulewidth}{\abovetopsep}{\belowbottomsep}

    \end{tabular}
    }
     \vspace{-18pt} 
\end{table}
\paragraph{Feature visualizations.} 
We evaluate \Pn{} by comparing its learned representations with VANILLA, FDS, and LDS. Using t-SNE visualization, we map ResNet-50's features to a 2D space for the AgeDB-DIR dataset. Fig.~\ref{fig: tsne} demonstrates the feature-label semantic correspondence exploited by \Pn{}, compared to VANILLA, FDS and LDS. 
VANILLA fails to effectively model the feature space regarding three key observations: a) high occurrences of collapses: features of minority samples are considerably collapsed to the majority ones. b) low relative spread: 
contradicting the linearity in the age estimation task's label space, the learned representation exhibits low feature variance across the label spread (across the colour spectrum) compared to the variance across a single label (same colour). c) Noticeable gaps within the feature space: contradicts the intended continuity in regression tasks. 
Compared to VANILLA, FDS and LDS slightly alleviate the semantic confusion in feature space. However, as shown in Fig.~\ref{tsne-d}, \Pn{} learns a considerably more effective feature space with fewer collapses, higher relative spread and semantic continuity.

\begin{figure}[t]
\vspace{-5mm}
  \centering
  
  \begin{subfigure}{0.16\textwidth}
    \includegraphics[width=\linewidth]{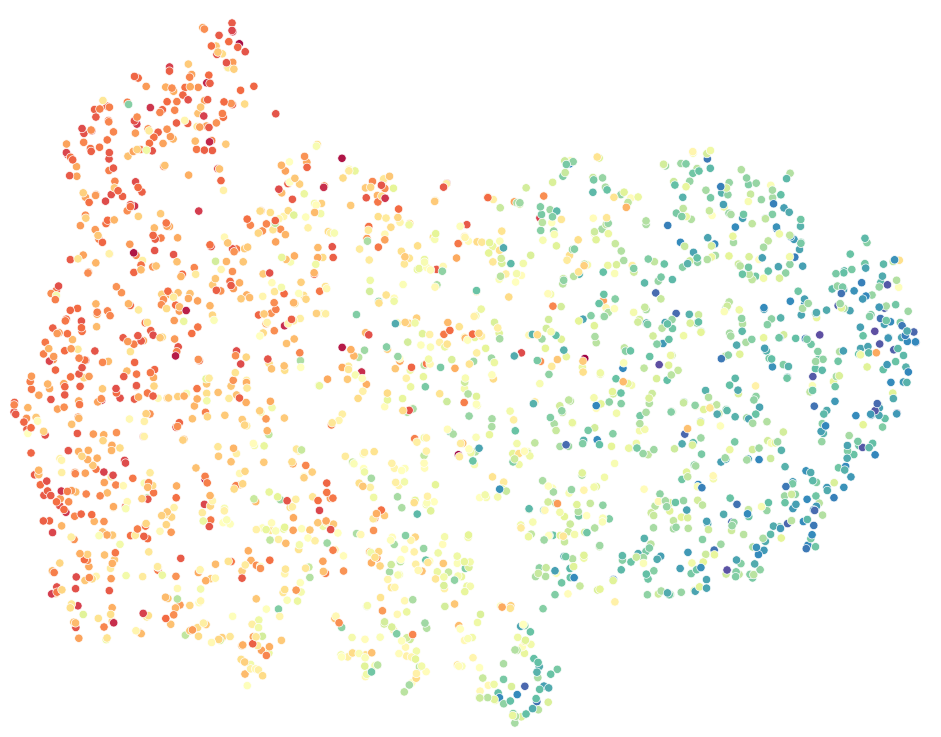}
    \caption{VANILLA}
  \end{subfigure}
  \hfil
  \begin{subfigure}{0.15\textwidth}
    \includegraphics[width=\linewidth]{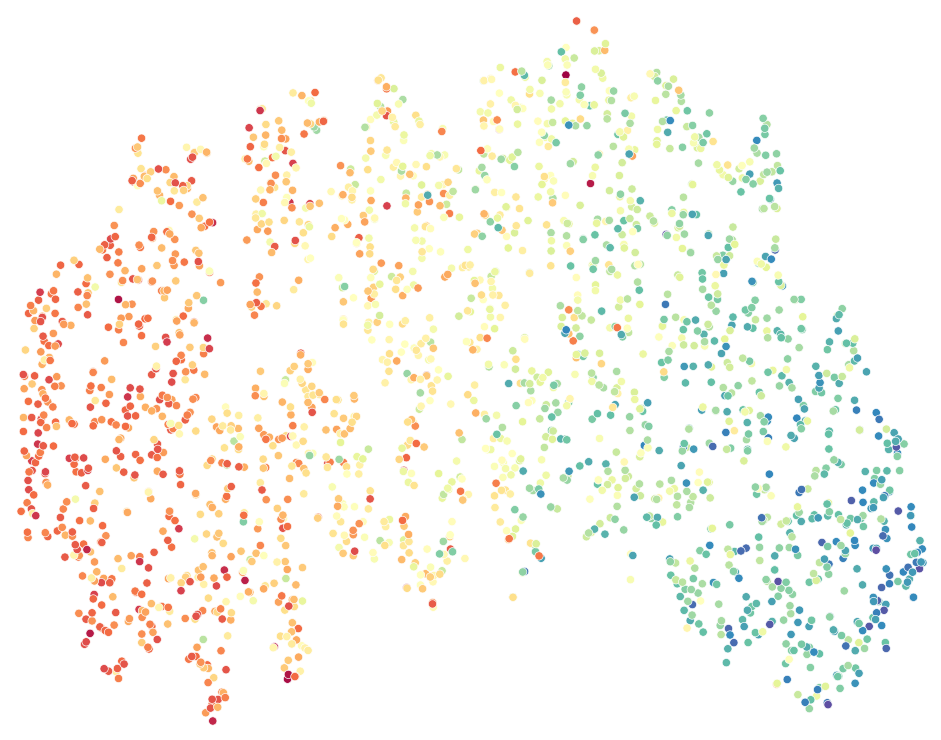}
    \caption{FDS}
  \end{subfigure}
  \hfil
  \begin{subfigure}{0.15\textwidth}
    \includegraphics[width=\linewidth]{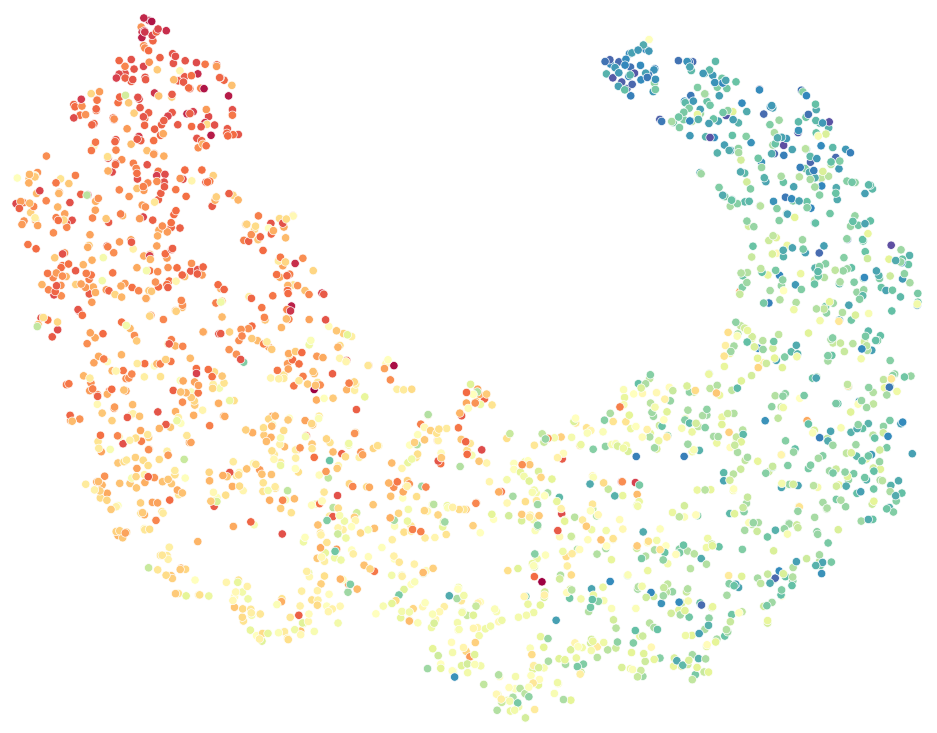}
    \caption{LDS}
  \end{subfigure}
  \hfil
  \begin{subfigure}{0.15\textwidth}
    \includegraphics[width=\linewidth]{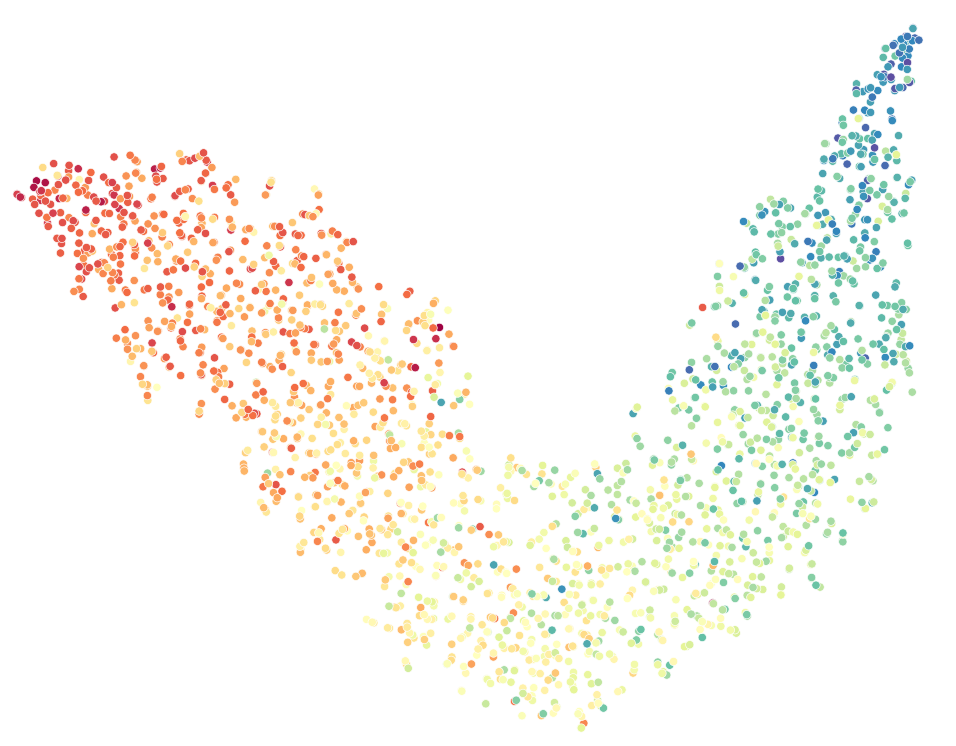}
    \caption{\Pn}
    \label{tsne-d}
  \end{subfigure}
  \begin{subfigure}{0.06\textwidth}
    \includegraphics[width=\linewidth]{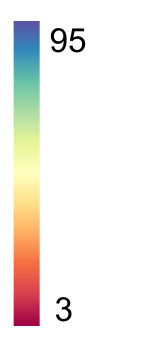}
    \vspace{1mm}
  \end{subfigure}
  \vspace{-8pt} 
  \caption{Feature visualization on AgeDB-DIR for (a)~VANILLA, (b)~FDS, (c)~LDS and (d)~\Pn{}. }
  \label{fig: tsne}
  \vspace{-18pt}
\end{figure}

\subsection{Ablation Studies on Design Modules} 
\vspace{-3mm}
\paragraph{Negative sampling, pushing weight and pushing power analysis.}Here we assess the significance of our method's contributions through the evaluation of several variations of \Pn{} and investigating the impact of different choices of the hyperparameters in \Pn.
We define four versions of \Pn{}: 
\textbf{Contrastive-\Pn}: contrastive regularizer where negative peers are selected only based on label similarities.
\textbf{\Pn{}-$\mathcal{S}$}: \Pn{} with \textbf{no} pushing power assigned to the negative pairs. \textbf{\Pn{}-$\eta$}: \Pn{} with pushing powers that \textbf{are not} proportionate to the instance weights. 
\textbf{\Pn{}-$Sim$}: \Pn{} with pushing powers that \textbf{are not} proportionate to the label similarities. Table~\ref{table: abl} describes the comparison between these three versions on the AgeDB-DIR benchmark and shows the crucial role of each component in deep imbalanced regression. \textbf{Contrastive-\Pn} is the continuous version of SupCon~\citep{supcon} that is biased to majority samples~\citep{kcl}. Thus, \textbf{Contrastive-\Pn} shows better results on \textit{many} shot that is due to the higher performance for majority samples, while it degrades the performance for minority ones. However, \Pn{} results in a more balanced performance with significant improvement for the minoirty shots.

\begin{table}[h]    
    \vspace{-5pt}
    \caption{\textbf{Ablation results on AgeDB-DIR benchmark.}}
    \vspace{-5pt}
    \centering
    \resizebox{0.55\textwidth}{!}{%
    
    \label{table: abl}
    \begin{tabular}{l?cccc}
    
        \specialrule{2\heavyrulewidth}{\abovetopsep}{\belowbottomsep}
        \textbf{Metrics}&\multicolumn{4}{c}{\textbf{MAE}$\downarrow$}\\
        \specialrule{1.5\heavyrulewidth}{\abovetopsep}{\belowbottomsep}
        \textbf{Methods/Shots} & \textbf{All} & \textbf{Many} & \textbf{Median} & \textbf{Few}\\
        \specialrule{1.5\heavyrulewidth}{\abovetopsep}{\belowbottomsep}
        \textbf{Contrastive-\Pn}&7.69&\textbf{6.12}&8.73&12.42\\
        \textbf{\Pn{}-$\mathcal{S}$}&7.51 & 6.49&8.23&10.86\\

        \textbf{\Pn{}-$Sim$}&7.54&6.43&8.19&10.69\\
        \textbf{\Pn{}-$\eta$}&7.52&6.55&8.11&10.51\\
        \Pn&\textbf{7.48}&6.53&\textbf{8.03}&\textbf{10.42}\\
        \specialrule{2\heavyrulewidth}{\abovetopsep}{\belowbottomsep}
        
    \end{tabular}
    }
     \vspace{-3mm}  
\end{table}
\paragraph{Similarity threshold analysis.} We investigate the choice of similarity threshold $\omega$ by exploring the learned features and model performance employing different values for the similarity threshold. Fig.~\ref{tsne-05} and Fig.~\ref{tsne-1} compare the feature space learnt with similarity threshold $\omega=2$ and $\omega=1$ for \Pn{} on the AgeDB-DIR benchmark. \Pn{} implicitly enforces feature smoothness and linearity in feature space. 
A high threshold ($\omega = 2$) is prone to encouraging feature smoothing in a limited label range and bias towards majority samples. As illustrated in Fig.~\ref{tsne-1}, choosing a \textcolor{blue}{lower} similarity threshold leads to smoother and more linear feature space. Fig.~\ref{abl} demonstrates the ablation results for the similarity threshold on the Age-DB-DIR dataset. For AgeDB-DIR, $\omega=1$ produces the best performance. An intuitive explanation could be the higher thresholds will impose sharing feature statistics to an extent that is not in correspondence with the heterogeneous dynamics in the feature space. For example, in the task of facial age estimation, the ageing patterns of teenagers are different from people in their 20s. For more details on this study please refer to Appendix~\ref{ablation appendix}.
\begin{figure}[ht]
  \centering
  \vspace{-3mm}
  \begin{subfigure}{0.27\textwidth}
    \includegraphics[width=0.6\linewidth]{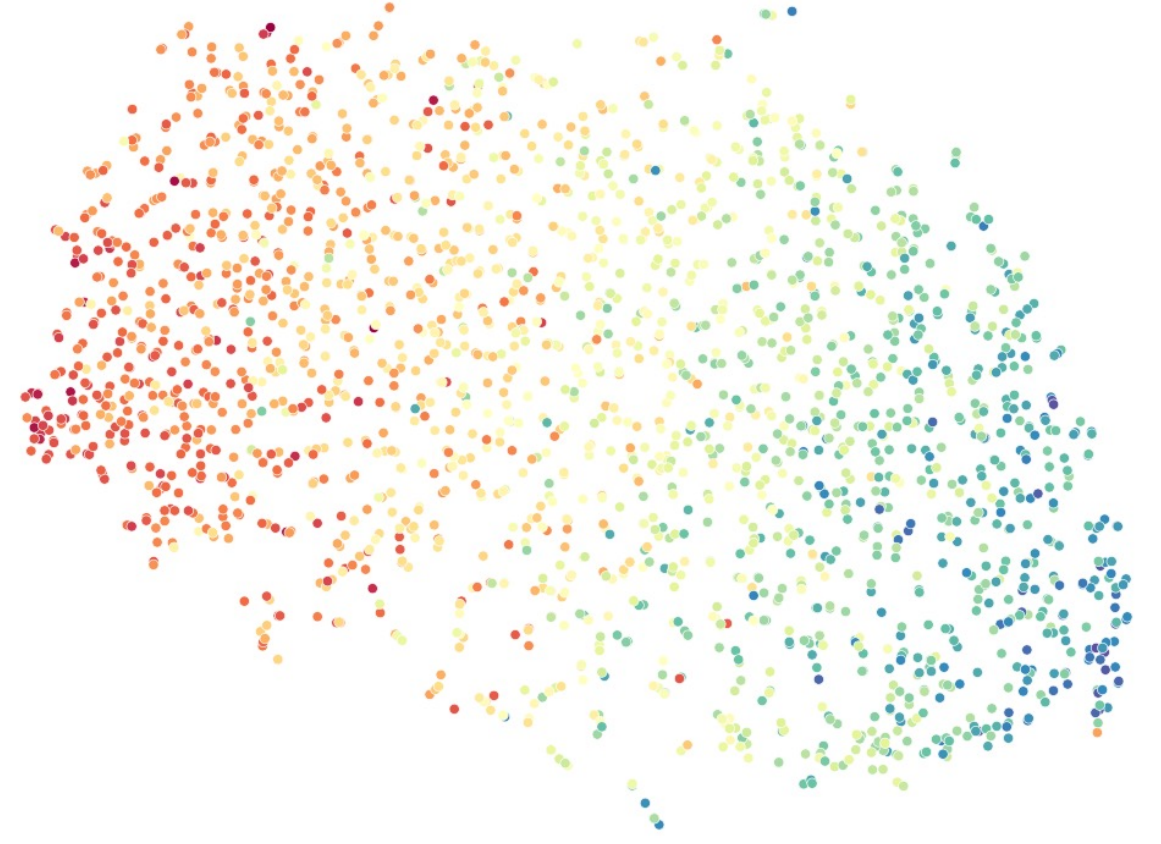}
    \caption{Visualization for $\frac{1}{\omega}=0.5$}
    \label{tsne-05}
  \end{subfigure}
  \hspace{-10mm}
  \begin{subfigure}{0.06\textwidth}    \includegraphics[width=\linewidth]{Figs/leg}
    \vspace{1mm}
  \end{subfigure}
  \begin{subfigure}{0.25\textwidth}
    \includegraphics[width=0.6\linewidth]{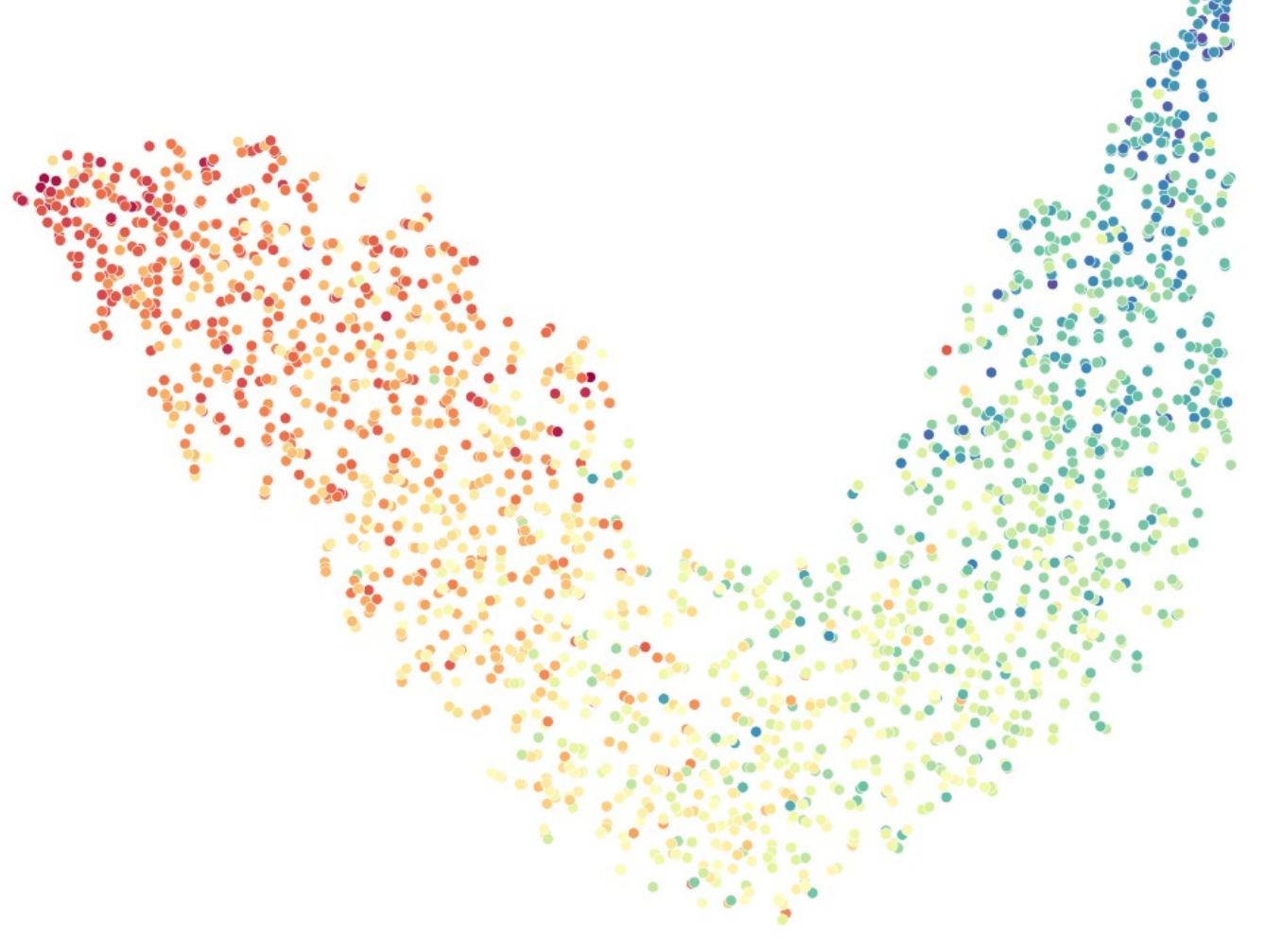}
    \caption{Visualization for $\frac{1}{\omega}=1$}
    \label{tsne-1}
  \end{subfigure}
  \hspace{-8mm}
  \begin{subfigure}{0.06\textwidth}    \includegraphics[width=\linewidth]{Figs/leg}
    \vspace{1mm}
    \label{w-a}
  \end{subfigure}
  \begin{subfigure}{0.30\textwidth}    \includegraphics[width=\linewidth]{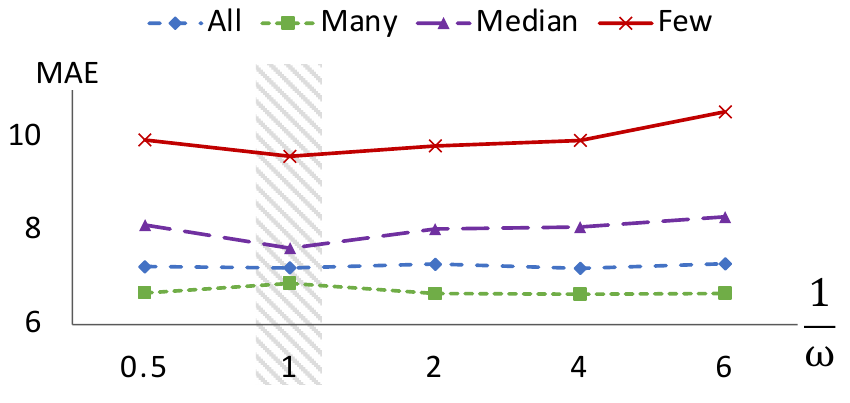}
    \caption{Similarity threshold analysis.}
    \label{abl}
  \end{subfigure}
  \vspace{-5pt}
  \caption{\textbf{Ablation study on the similarity threshold $\omega$}. (a) and (b) compares the learnt feature space for similarity threshold of $2$ and $1$ respectively. (c) Comparison of different choices of $\omega$ in terms of MAE.}
  \label{fig: tsne-w}
  \vspace{-8mm}
\end{figure}
\section{Conclusion}
\vspace{-3mm}
In this work, we propose \Pn{}, a novel regularizer for DIR that incorporates continuity to contrastive learning and implicitly encourages preserving local and global semantic relations in the feature space without assumptions about inter-label dependencies.
The novel anchor selection proposed in this work consistently derives a balanced training focus across the imbalanced training distribution. \Pn{} is orthogonal to all regression models. 
Our experiments on uni- and multi-dimensional DIR benchmarks show that regularizing a regression model with \Pn{} considerably lifts its performance, especially for the under-represented samples and high-dimensional label spaces. \Pn{} opens a new perspective to contrastive regression on imbalanced data.

\bibliography{iclr2024_conference}
\bibliographystyle{iclr2024_conference}

\appendix
\newpage
\appendix
\section{Experiments}

\subsection{Empirical analysis of the motivation}
\label{Empirical analysis of the motivation}

The key motivation of the design of \Pn{} is that features of minority samples tend to collapse to their majority neighbours. \Pn{} highlights these situations by defining a penalty based on the misalignments between prediction similarities and label similarities. Further, \Pn{} regularizes a regression model by minimizing the defined penalty in a contrastive manner.

Here we first define the penalty $P(\mathrm{y})$ for each prediction value $\mathrm{y} \in \mathcal{D}_y$. $P(\mathrm{y})$ denotes the average of the regression errors of the samples with the similar prediction value of $\mathrm{y}$: $P(\mathrm{y}) = \frac{1}{N_{\mathrm{y}}}\sum^{N_{\mathrm{y}}}_{i=0}\mathcal{L}_{\mathcal{R}}(x_i,y_i)$, where $\mathcal{R}(\mathcal{E}(x_i)) \simeq \mathrm{y}$, $y_i \not\simeq \mathrm{y}$ and $N_\mathrm{y}$ is the number of the  samples collapsed to the feature space of samples labelled with $\mathrm{y}$.  


To empirically confirm the motivation behind \Pn{}, we investigate the importance of the penalty term in regression tasks. In addition, we show that regularizing a regression model with \Pn{} consistently alleviate the penalty term defined by \Pn{} and this optimization considerably contributes to imbalanced regression. 
Fig.~\ref{fig: reg_t} and Fig.~\ref{fig: reg_v} demonstrate the comparison between LDS and \Pn{} in terms of the training loss curve and validation loss curve, respectively. Moreover, Fig.~\ref{fig: reg_p} shows the trend of the expected value of $P(\mathrm{y})$ over the label space, throughout the training. Comparing Fig.~\ref{fig: reg_p} with Fig.~\ref{fig: reg_t} and Fig.~\ref{fig: reg_v}, $P(\mathrm{y})$ follows the same decreasing pattern as training loss and validation loss. This observation show that penalizing the penalty term is highly coupled with the learning process. In addition, Fig.~\ref{fig: reg} shows that \Pn{} outperform LDS with a considerable gap, particularly in terms of $P(\mathrm{y})$; showing that \Pn{} regularizer consistently alleviates the penalty and significantly contributes to the imbalanced regression.
\begin{figure}[hbt!]
  \centering
  \begin{subfigure}{0.32\textwidth}    \includegraphics[width=\linewidth]{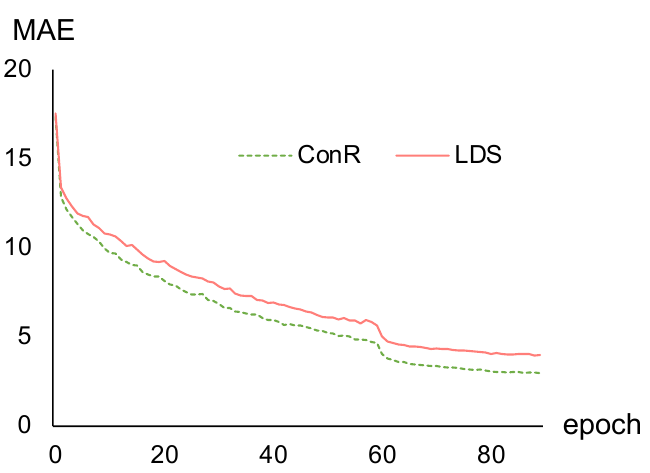}
    \caption{Training loss}
    \label{fig: reg_t}
  \end{subfigure}
  \begin{subfigure}{0.32\textwidth}    \includegraphics[width=\linewidth]{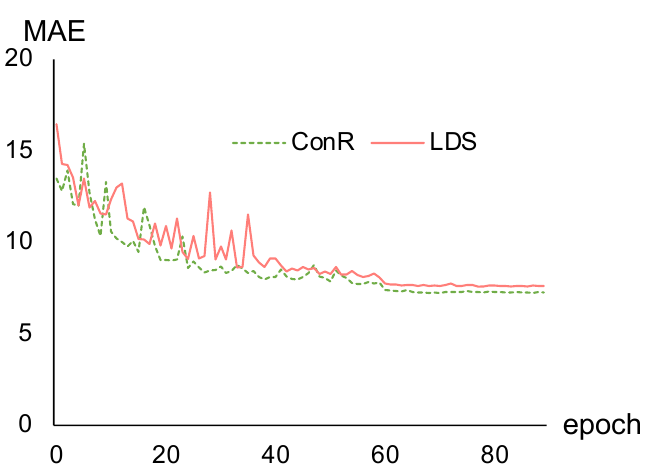}
    \caption{Validation loss}
    \label{fig: reg_v}
  \end{subfigure}
  \begin{subfigure}{0.32\textwidth}    \includegraphics[width=\linewidth]{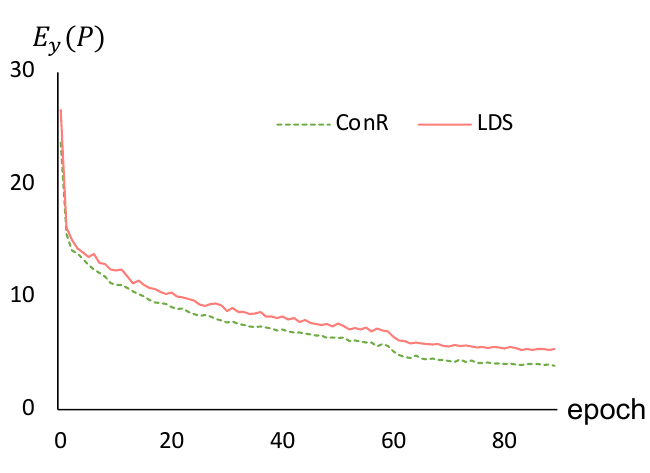}
    \caption{Expected value of penalty}
    \label{fig: reg_p}
  \end{subfigure}
  
  \caption{Comparison of the performance of LDS and \Pn{} on  AgeDB-DIR benchmark in terms of (a)training loss, (b)validation loss, and (c)regularizer penalty.}
  \label{fig: reg}
\end{figure}

Fig.~\ref{fig: diff-l} shows the training label distribution and Fig.~\ref{fig: diff-p} depicts the difference of $P(\mathrm{y})$ across the label space between \Pn{} and LDS. It empirically shows the considerable improvement of \Pn{} over LDS in terms of the defined penalty. More improvement over the majority of samples is intuitive because due to the imbalanced distribution, most of the  collapses in feature space happen in the majority areas and decreasing the penalty in these areas contributes the most to the imbalanced regression. 
Finally, Fig.~\ref{fig: diff-e} compares the regression error of \Pn{} and LDS and we observe \Pn{} results in a considerable improvement over LDS, especially for minority samples. 
\begin{figure}[hbt!]
  \centering
  \begin{subfigure}{0.32\textwidth}    \includegraphics[width=\linewidth]{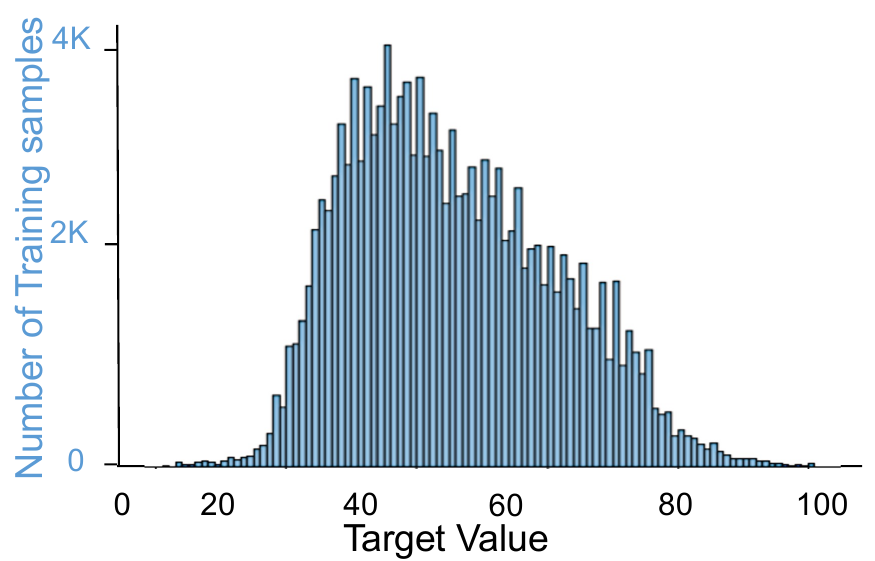}
    \caption{Training loss}
    \label{fig: diff-l}
  \end{subfigure}
  \begin{subfigure}{0.32\textwidth}    \includegraphics[width=\linewidth]{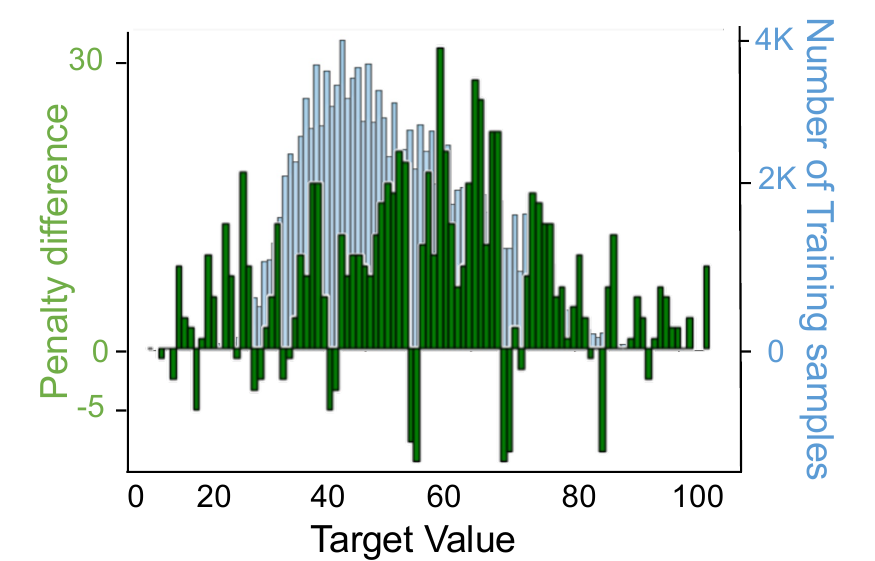}
    \caption{Validation loss}
    \label{fig: diff-p}
  \end{subfigure}
  \begin{subfigure}{0.32\textwidth}    \includegraphics[width=\linewidth]{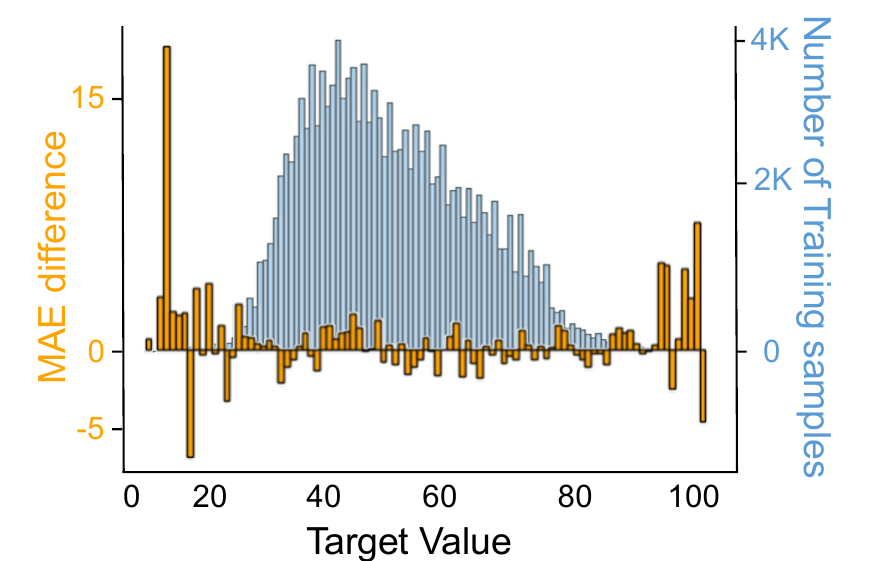}
    \caption{Expected value of penalty}
    \label{fig: diff-e}
  \end{subfigure}
  
  \caption{Qualitative analysis of \Pn{} in terms of regularizer penalty on AgeDB-DIR benchmark. (a) training label distribution. (b) Penalty difference (LDS minus \Pn) on balanced test data. (c) MAE/ difference (LDS minus \Pn) on balanced test data. }
  \label{fig: diff}
\end{figure}
\subsection{Dataset details}
\label{dataset details}
\paragraph{Age estimation.}
We evaluated our method on two DIR benchmarks for age estimation curated by~\cite{yang2021delving}: IMDB-WIKI-DIR and AgeDB-DIR.
IMDB-WIKI~\citep{rothe2018deep} has 191.5K images for training, and 11.0K images for validation and testing, respectively. To structure IMDB-WIKI-DIR, ~\citet{yang2021delving} bin the label space with a bin length of 1 year,  where the minimum age is 0 and the maximum age is 186. The bin density varies between 1 and 7,149. 
The AgeDB dataset~\citep{moschoglou2017agedb} has  16,488 samples.\citet{yang2021delving} constructed AgeDB-DIR in a similar manner as IMDB-WIKI-DIR, where the minimum age is 0 and the maximum age is 101. The maximum number of samples per bin is 353 images and the minimum bin density is 1. There are 12,208 training samples and the validation set and test set are made balanced with 2,140 samples each.
\paragraph{Depth estimation.}
\citet{yang2021delving} created NYUD2-DIR based on the NYU Depth Dataset V2~\cite{silberman2012indoor}. NYU Depth Dataset V2 has images and corresponding depth maps for different indoor scenes and the task is to predict the depth maps from the RGB scene images.  The upper bound of the depth maps is 10 meters and the lower bound is 0.7 meters. Following standard practices. There are 50K images for training and 654 images for testing. \citet{yang2021delving} use the bin length of 0.1-meter bin density varies between 1.13 × 106 and 1.46 × 108. For a balanced test set, \citet{yang2021delving} randomly select 9,357 test pixels for each bin from 654 test images with a total of 8.70 × 105 test pixels in the NYUD2-DIR test set.
\paragraph{Gaze estimation.} We used a subset of the MPIIGaze dataset comprising 45,000 training samples from 15 individuals, with 3,000 samples per person. The dataset is naturally imbalanced over the 2D training label distribution.

Here we provide the details of deriving the proposed MPIIGaze-DIR benchmark from the MPIIGaze dataset. the label space of MPIIGaze is 2-dimensional with one dimension ranging from -0.39 to 0.08 and the other from -0.72 to 0.67. Fig.~\ref{fig: gazedist} shows the imbalanced distribution across each dimension separately and  Fig.~\ref{fig: gazehist} shows the joint distribution of the label space with a grid size of 10. Considering the joint distribution, we define thresholds on the bin densities to curate shots~(i.e. \textit{many}, \textit{median}, and \textit{few}) as follows: 1300 or more for the \textit{many}-shot, 700 to 1300 for the \textit{median}-shot and less than 700 for the \textit{few}-shot. Next, with the code snippet provided in Fig.~\ref{fig: gazecode}, We assign the samples to their corresponding shots.

\begin{figure}[hbt!]
  \centering
  \begin{subfigure}{0.32\textwidth}    \includegraphics[width=\linewidth]{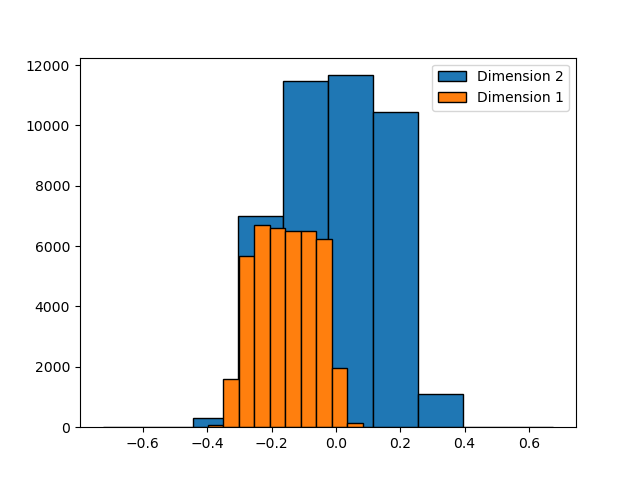}
    \caption{Individual distribution}
    \label{fig: gazedist}
  \end{subfigure}
  \begin{subfigure}{0.32\textwidth}    \includegraphics[width=\linewidth]{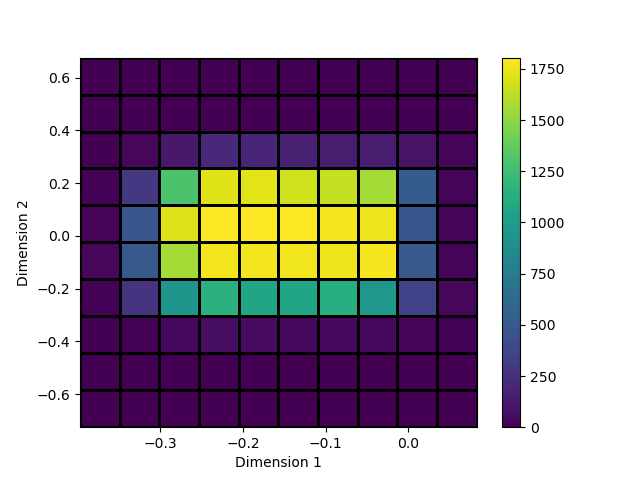}
    \caption{Joint distribution}
    \label{fig: gazehist}
  \end{subfigure}
  \caption{Distribution of the 2-dimensional label space of MPIIGaze benchmark.}
  
\end{figure}

\begin{figure}[hbt!]
  \centering
  \includegraphics[width=0.6\linewidth]{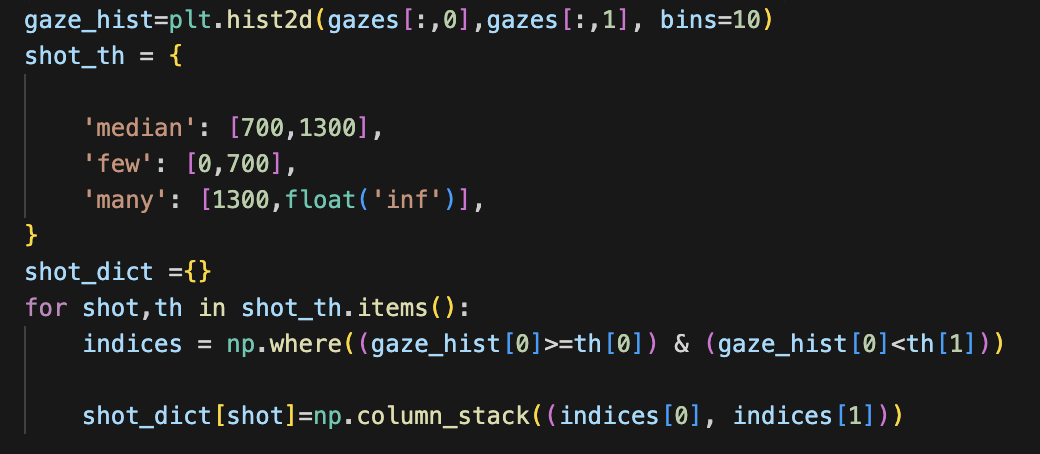}
  \caption{Code snippet for MPIIGaze-DIR creation.}
  \label{fig: gazecode}
\end{figure}

For training, we follow a leave-one-out scheme where each time one person is dedicated to validation, one person for testing and 13 people for training. The reported results are in terms of average test results among all the 15 people. To create a balanced test set, we take 200 samples from each shot
similar to the methods used in the FDS work. 
Follow the baselines \citep{ranksim}, only the training data for these tasks is imbalanced; the test dataset is balanced.

\subsection{Baselines}
\label{appendix: baselines}
\Pn{} is orthogonal to state-of-the-art imbalanced learning methods, thus we examine the improvement from \Pn{} when added on top of existing methods, which we refer to as baselines for our technique: 
Label- and Feature-Distribution Smoothing (LDS and FDS) encourage local similarities in label and feature space~\citep{yang2021delving}. 
RankSim imposes a matching between the order of similarities in label space with these similarities in feature space~\citep{ranksim}. Balanced MSE encourages a balanced prediction distribution~\citep{ren2022balanced}.
To investigate the effect of contrastive learning on deep imbalanced regression, we also regularize using infoNCE~\citep{oord2018representation} and contrastive architecture of MoCo~\citep{moco}, MoCo V2~\citep{chen2020improved}. Refer to Appendix~\ref{ablation appendix} for contrastive analysis. 
Results for RankSim on depth estimation and gaze estimation are omitted as RankSim is not suitable for these tasks.

\subsection{Implementation details}
\label{implementation details}

We use four NVIDIA GeForce GTX 1080 Ti GPU to train all models. For a fair comparison, we follow \citep{yang2021delving} for all standard train/val/test splits. The rest of this section provides the implementation details and choices of hyperparameters for all three datasets.

\paragraph{Age estimation.}
For AgeDB-DIR benchmark and IMDB-WIKI-DIR benchmark, we use Resnet50  for encoder $\mathcal{E}(\cdot)$ and a one-layer fully connected network for the regression module $\mathcal{R}(\cdot)$. The batch size is 64 and the learning rate is  $2.5*10^{-4}$ and decreases by 10$\times$ at epoch 60 and epoch 80. We use the Adam optimizer with a momentum of 0.9 and a weight decay of 1e-4. Following the baselines~\citep{yang2021delving} the loss function for regression $\mathcal{L}_{R}$ is Mean Absolute Error(MAE). All the models are trained for 90 epochs. The augmentations in the age estimation task are random crop and random horizontal flip.

$\eta_j =(0.01)w_j$ and the similarity function $Sim(\cdot, \cdot)$ is inverse Mean Absolute Error(MAE). To resolve \textit{divide by zero} and \textit{infinite numbers}, a pair of samples with MAE distance  $<\frac{1}{\omega}$ are considered similar. 
Further, the pushing weight $\mathcal{S}_{j,q}$ is defined as: $\mathcal{S}_{j,q} = f_\mathcal{S}(\eta_j,\frac{1}{{MAE}(y_j,y_q)}) = \eta_j {MAE}(y_j,y_q)$.
Finally, the similarity threshold $\omega$ is 1, $\tau =0.2$, $\alpha=1$, and $\beta =4$.


\paragraph{Depth estimation.}
For NYUD2-DIR benchmark, we use ResNet-50-based encoder-decoder architecture~\citep{hu2019revisiting}. The output size is 114 × 152. The batch size is 32 and the learning rate is $1*10^{-4}$. All models are trained for 90 epochs with an Adam optimizer. The momentum of the optimizer is 0.9 and its weight decay is $1e-4$. Following the baselines~\citep{yang2021delving,hu2019revisiting} the loss function for regression $\mathcal{L}_{R}$ is root-mean-square(RMSE). The augmentations in the depth estimation task are random rotate, color jitter, and random horizontal flip.

To measure the similarity in label space, first, for each sample $(z_j,y_j)$, we take the average value of the depth map, denoted as $\bar{y_j}$. Then, for each pair of samples, we use the root-mean-square(RMSE) to quantify the similarity of this pair in the label space. If the RMSE$(\bar{y_i},\bar{y_j})$ is less than $\frac{1}{\omega}$, sample $i$ and sample $j$ are considered similar and otherwise dissimilar.
In addition, the pushing weight $\mathcal{S}_{j,q}$ is defined as: $\mathcal{S}_{j,q} = f_\mathcal{S}(\eta_j,\frac{1}{{RMSE}(y_j,y_q)}) = \eta_j {RMSE}(y_j,y_q)$. 
The similarity threshold $\omega$ is 5, $\beta=0.2$, $\tau =0.7$ and $\eta_j =(0.2)w_j$.
\paragraph{Gaze estimation.}
 We reported our results on a balanced test set, containing 600 samples in total and 200 samples per each "many," "median," and "few" shots. Our results were averaged over five random runs to ensure statistical significance. Each run in the evaluation incorporated a leave-one-out scheme, where we performed 15 runs with a single individual as the designated test set. The final results are the Mean Angle Error (in degrees) for all the individuals. The backbone is LeNet, $\beta = 0.4$, $\alpha = 1$, $\omega = 1$ 
. $\eta_j =(0.01)w_j$ and $\mathcal{S}_{j,q} = f_\mathcal{S}(\eta_j,\frac{1}{{MAE}(y_j,y_q)}) = \eta_j {MAE}(y_j,y_q)$.
The batch size and the base learning rate are 32 and 0.01, respectively.
The augmentations in the gaze estimation task are random crop, random resize, and colour jitter.

\subsection{Performance Analysis}
\label{Performance Analysis}
All the experimental results are reported as the averages of 5 random runs.

\paragraph{More results for age estimation.}
For a more extensive empirical confrimation that \Pn{} is orthogonal to DIR baselines, Table~\ref{table: age-more} shows the performance improvements when RRT~\cite{yang2021delving}, Focal-R~\cite{yang2021delving} and Balanced MSE are regularized by \Pn{}.
\begin{table}[htb!]    
    \caption{\textbf{MAE results of \Pn{} on AgeDB-DIR Benchmark and IMDB-WIKI-DIR benchmark.}}
    \resizebox{\textwidth}{!}{%
    \centering
    \label{table: age-more}
    \begin{tabular}{l?cccc:cccc}
    
        \specialrule{2\heavyrulewidth}{\abovetopsep}{\belowbottomsep}
        \textbf{Metric}&\multicolumn{8}{c}{\textbf{MAE}$\downarrow$}\\
        
        \specialrule{1.5\heavyrulewidth}{\abovetopsep}{\belowbottomsep}
        \textbf{Benchmark}&\multicolumn{4}{c:}{\textbf{AgeDB-DIR}}&\multicolumn{4}{c}{\textbf{IMDB-WIKI-DIR}}\\
        \specialrule{1.5\heavyrulewidth}{\abovetopsep}{\belowbottomsep}
        
        \textbf{Methods/Shots} & \textbf{All} & \textbf{Many} & \textbf{Median} & \textbf{Few} & \textbf{All} & \textbf{Many} & \textbf{Median} & \textbf{Few}\\
        \specialrule{1.5\heavyrulewidth}{\abovetopsep}{\belowbottomsep}
        RRT &7.74	&6.98	&8.79&	11.99&7.81	&7.07	&14.06&	25.13\\
        RRT + \Pn{} (\textbf{Ours})&\textbf{7.53} &	\textbf{6.79}&	\textbf{7.60}&	\textbf{10.30} &\textbf{7.41}	&\textbf{6.89}&	\textbf{13.20}&	\textbf{23.30}\\
        \hline
        Focal-R &7.64 &	6.68&	9.22&	13.00& 7.97	&7.12	&15.14	&26.96\\
        Focal-R + \Pn{} (\textbf{Ours})&\textbf{7.23}	 &\textbf{6.63}&	\textbf{8.30}	&\textbf{11.89} &\textbf{7.85}&	\textbf{7.01}	&\textbf{14.31}&\textbf{25.23}\\
        \hline
        Balabced MSE~(GAI) &7.57	&7.46&	8.40	&10.93& 8.12	&7.58&	12.27&	23.05\\
       
        Balabced MSE~(GAI) + \Pn{} (\textbf{Ours})&\textbf{7.22}	&\textbf{6.71}&	\textbf{7.99}	&\textbf{9.88} &\textbf{7.84}&	\textbf{7.20}	&\textbf{12.09}&	\textbf{22.20}\\
        \specialrule{1.5\heavyrulewidth}{\abovetopsep}{\belowbottomsep}
        \textbf{Ours} vs. RRT & \textcolor{green}{2.71 \%}   & \textcolor{green}{2.72\%}   & \textcolor{green}{13.54\%}   & \textcolor{green}{14.10 \%}&\textcolor{green}{5.12 \%}   & \textcolor{green}{2.55\%}   & \textcolor{green}{6.12\%}   & \textcolor{green}{7.28 \%}\\
        \textbf{Ours} vs. Focal-R & \textcolor{green}{5.37 \%}   & \textcolor{green}{0.75\%}   & \textcolor{green}{9.98\%}   & \textcolor{green}{8.54 \%}&\textcolor{green}{1.51 \%}   & \textcolor{green}{1.54\%}   & \textcolor{green}{5.48\%}   & \textcolor{green}{6.42 \%}\\
        \textbf{Ours} vs. Balanced MSE~(GAI) & \textcolor{green}{4.62 \%}   & \textcolor{green}{10.05\%}   & \textcolor{green}{4.88\%}   & \textcolor{green}{9.61 \%}&\textcolor{green}{3.45 \%}   & \textcolor{green}{5.01\%}   & \textcolor{green}{1.47\%}   & \textcolor{green}{3.69 \%}\\
        \specialrule{2\heavyrulewidth}{\abovetopsep}{\belowbottomsep}
        
    \end{tabular}
    }
\end{table}
\paragraph{Error Reduction.}
Here we show the comparison of the Error reduction resulting from adding \Pn{} to the deep imbalanced regression baselines (LDS, FDS and RankSim) for age estimation benchmarks(e.g. AgeDB-DIR and IMDB-WIKI-DIR) and (LDS, FDS and Balanced MSE) for gaze estimation benchmark. Fig.~\ref{fig: p-a}, Fig.~\ref{fig: p-i} and Fig.~\ref{fig: p-g} empirically confirm significant performance consistency \Pn{} introduces to DIR for  AgeDB-DIR, IMDB-WIKI-DIR and MPIIGaze-DIR benchmarks, respectively.
\begin{figure}[htb!]
  \centering
  \begin{subfigure}{0.32\textwidth}
    \includegraphics[width=1\linewidth,height =32mm]{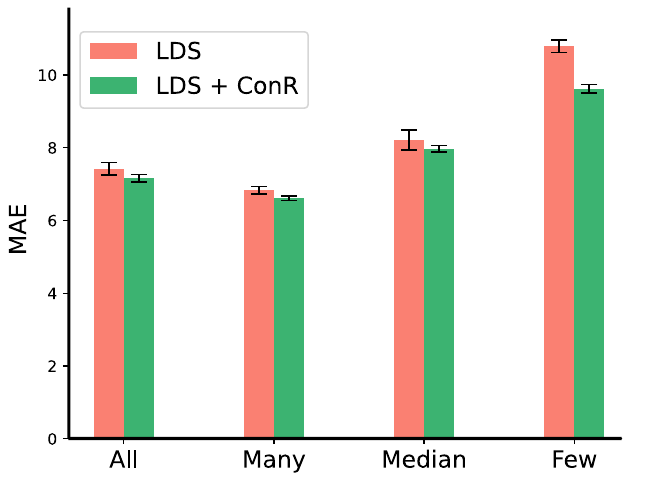}
  \end{subfigure}
  \hfill
  \begin{subfigure}{0.32\textwidth}   \includegraphics[width=1\linewidth,height =32mm]{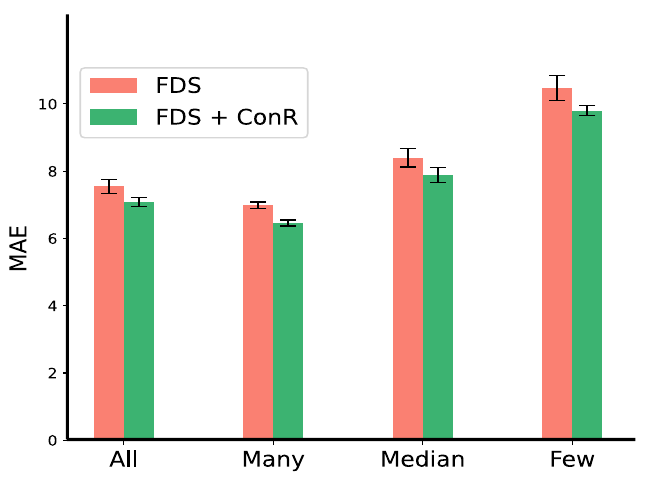}
  \end{subfigure}
  \hfill
  \begin{subfigure}{0.32\textwidth}  \includegraphics[width=1\linewidth,height =32mm]{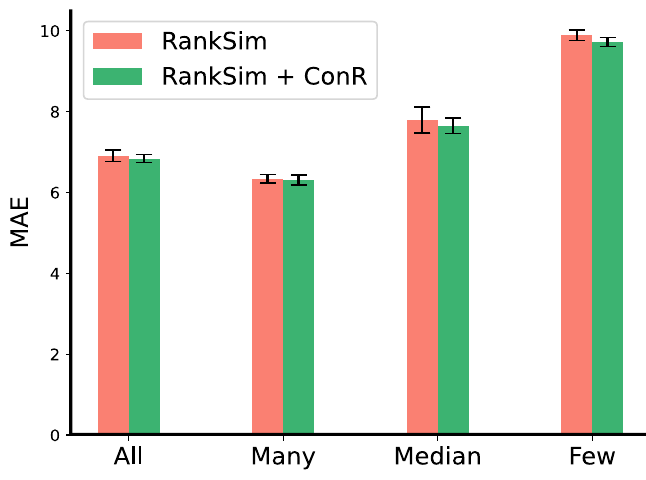}
  \end{subfigure}
  \caption{comparison on MAE results by adding \Pn{} on top of the baselines for AgeDB-DIR benchmark.}
  \label{fig: p-a}
\end{figure}

\begin{figure}[htb!]
  \centering
  \begin{subfigure}{0.32\textwidth}
    \includegraphics[width=1\linewidth,height =32mm]{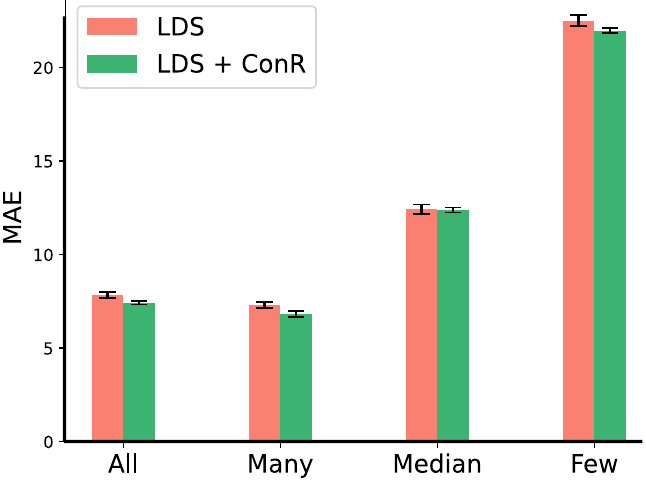}
  \end{subfigure}
  \hfill
  \begin{subfigure}{0.32\textwidth}   \includegraphics[width=1\linewidth,height =32mm]{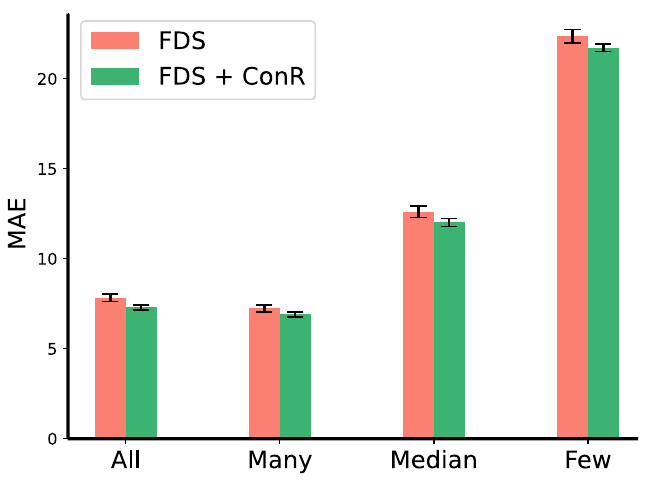}
  \end{subfigure}
  \hfill
  \begin{subfigure}{0.32\textwidth}  \includegraphics[width=1\linewidth,height =32mm]{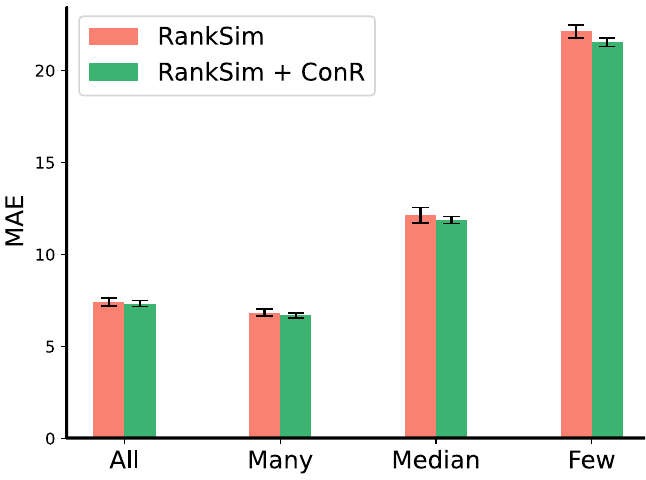}
  \end{subfigure}
  \caption{Comparison of the performance gain by regularizing the DIR baselines (LDS, FDS, RankSim) with \Pn{} on IMDB-WIKI-DIR benchmark.}
  \label{fig: p-i}
\end{figure}

\begin{figure}[htb!]
  \centering
  \begin{subfigure}{0.32\textwidth}
    \includegraphics[width=1\linewidth,height =32mm]{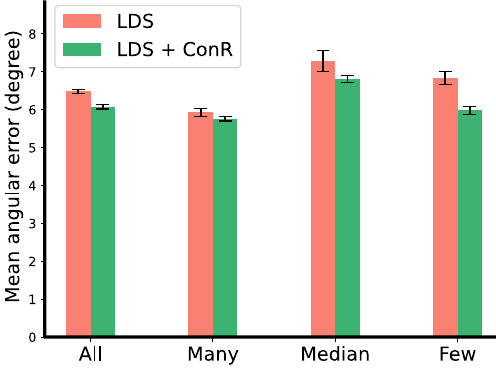}
  \end{subfigure}
  \hfill
  \begin{subfigure}{0.32\textwidth}   \includegraphics[width=1\linewidth,height =32mm]{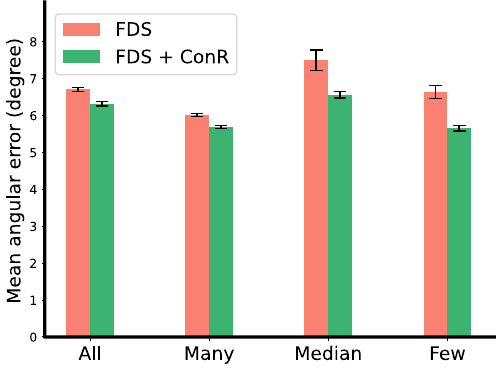}
  \end{subfigure}
  \hfill
  \begin{subfigure}{0.32\textwidth}  \includegraphics[width=1\linewidth,height =32mm]{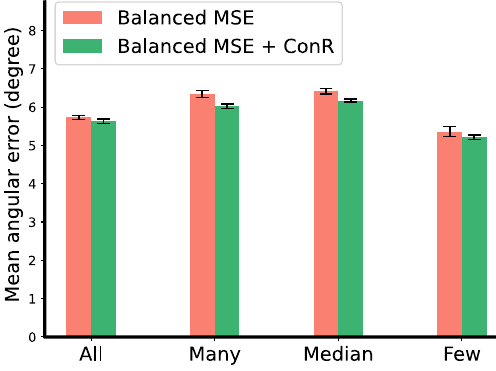}
  \end{subfigure}
  \caption{Comparison of the performance gain by regularizing the DIR baselines (LDS, FDS, Balanced MSE) with \Pn{} on MPIIGaze-DIR benchmark.}
  \label{fig: p-g}
\end{figure}

\paragraph{Feature visualization.} Fig.~\ref{Fig:tsne-2} compares the learned representations by RankSim and Balanced MSE. RanKSim by imposing order relationships encourage high relative spread while Balanced MSE suffer from low relative spread. Both RankSim and Balanced MSE have high occurrences of collapses and noticeable gaps in their feature space.  Comparing Fig.~\ref{fig: tsne}-d with Fig.~\ref{Fig:tsne-2} shows that \Pn{} learns the most effective representations.
\begin{figure}[t]
  \centering
  
  \begin{subfigure}{0.16\textwidth}
    \includegraphics[width=\linewidth]{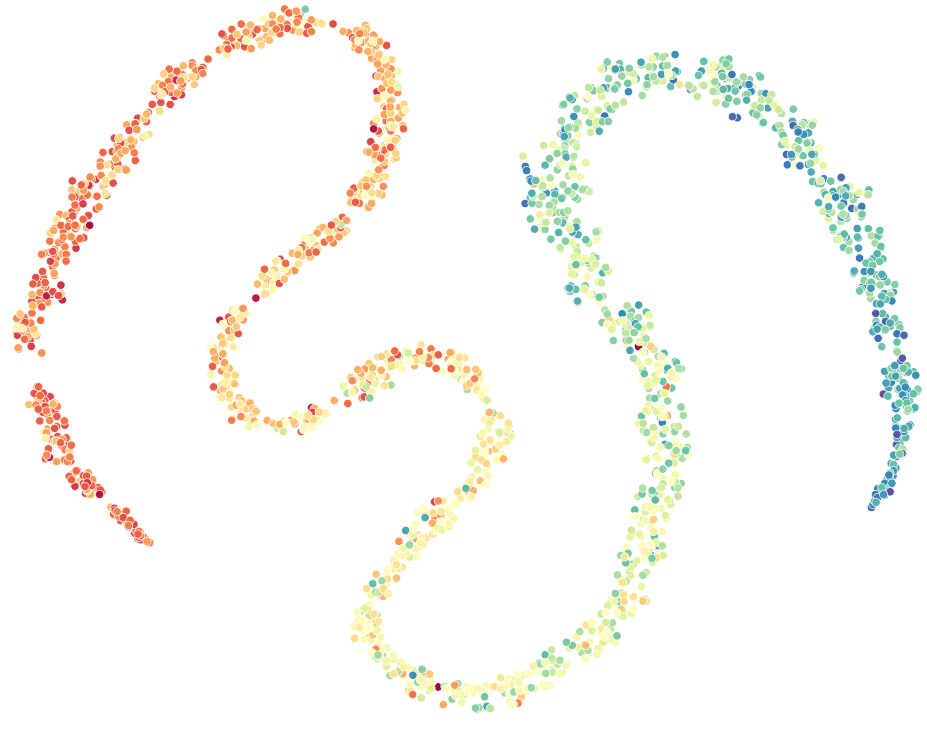}
    \caption{RankSim}
  \end{subfigure}
  \hfil
  \begin{subfigure}{0.15\textwidth}
    \includegraphics[width=\linewidth]{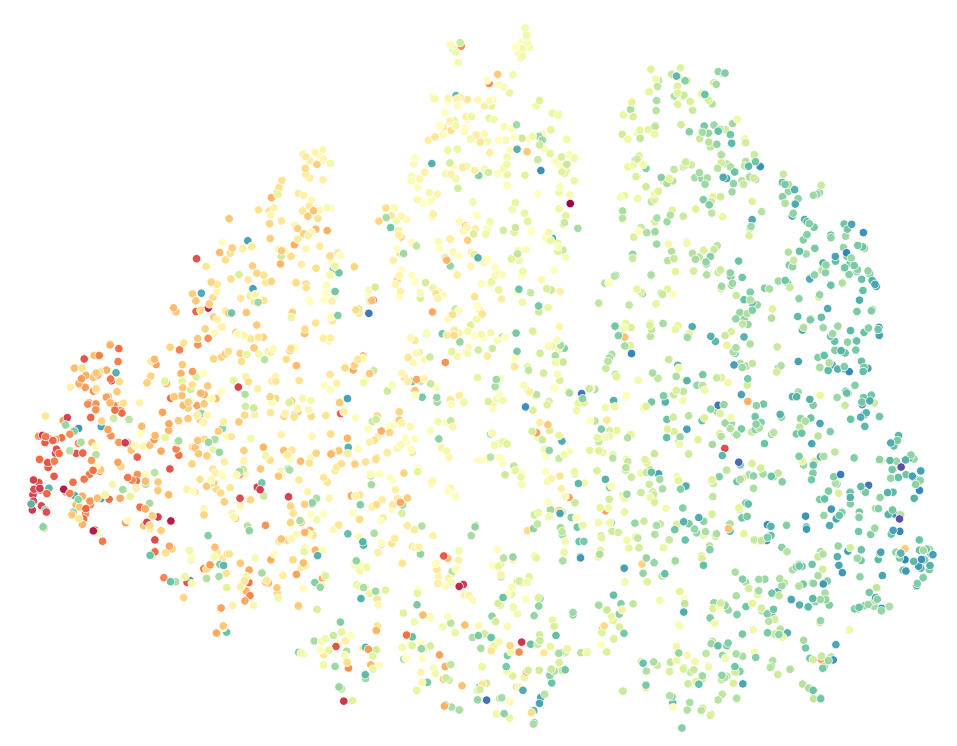}
    \caption{Balanced MSE}
  \end{subfigure}
  \caption{Feature visualization on AgeDB-DIR dataset for (a)~RankSim, (b)~Balanced MSE.}
  \label{Fig:tsne-2}
\end{figure}
\color{black}

\subsection{Ablation Study}
\label{ablation appendix}
\paragraph{Negative Sampling, Pushing Weight and Power Analysis.} Table~\ref{table: abl-i}, table~\ref{table: abl-n} and table~\ref{table: abl-g} show the significance of the main contributions of \Pn{} for IMDB-WIKI-DIR, NYUD2-DIR and MPIIGaze-DIR benchmarks, respectively.
\begin{table}[h]    
    \caption{\textbf{Ablation results of the design modules of \Pn{} on IMDB-WIKI-DIR benchmark.}}
    \centering
    \label{table: abl-i}
    \begin{tabular}{l?cccc}
    
        \specialrule{2\heavyrulewidth}{\abovetopsep}{\belowbottomsep}
        \textbf{Metric}&\multicolumn{4}{c}{\textbf{MAE}$\downarrow$}\\
        \specialrule{1.5\heavyrulewidth}{\abovetopsep}{\belowbottomsep}
        \textbf{Method/Shot} & \textbf{All} & \textbf{Many} & \textbf{Median} & \textbf{Few} \\
        \specialrule{1.5\heavyrulewidth}{\abovetopsep}{\belowbottomsep}
       
        \textbf{Contrastive-\Pn}&8.10&\textbf{6.79}&15.87&26.51\\
        \textbf{\Pn{}-$\mathcal{S}$}&7.79 & 6.99&14.61&25.64\\

        \textbf{\Pn{}-$Sim$}&7.83&6.87&14.30&25.59\\
        \textbf{\Pn{}-$\eta$}&\textbf{7.76}&7.10&14.25&25.33\\
        \Pn~\textbf{(Ours)} &7.84	&7.09	&\textbf{14.16}&	\textbf{25.15}\\
        \specialrule{2\heavyrulewidth}{\abovetopsep}{\belowbottomsep}
        
    \end{tabular}
\end{table}
\begin{table}[hbt!] 
\caption{\textbf{Ablation results of the design modules of \Pn{} on NYUD2-DIR benchmark.}}
    \centering
    \label{table: abl-n}
    \begin{tabular}{l?cccc}
    
        \specialrule{2\heavyrulewidth}{\abovetopsep}{\belowbottomsep}
        \textbf{Metric}&\multicolumn{4}{c}{\textbf{RMSE}$\downarrow$}\\
        \specialrule{1.5\heavyrulewidth}{\abovetopsep}{\belowbottomsep}
        \textbf{Method/Shot} & \textbf{All} & \textbf{Many} & \textbf{Median} & \textbf{Few} \\
        \specialrule{1.5\heavyrulewidth}{\abovetopsep}{\belowbottomsep}
        \textbf{Contrastive-\Pn}&1.518&\textbf{0.586}&1.124&2.412\\
        \textbf{\Pn{}-$\mathcal{S}$}&1.410 & 0.670&0.941&1.954\\

        \textbf{\Pn{}-$Sim$}&1.383&0.667&0.935&1.929\\
        \textbf{\Pn{}-$\eta$}&1.318&0.693&0.892&1.910\\
        \Pn{}~\textbf{(Ours)}&\textbf{1.304}&0.682&\textbf{0.889}&\textbf{1.885}\\
        \specialrule{2\heavyrulewidth}{\abovetopsep}{\belowbottomsep}
        
    \end{tabular}
\end{table}

\begin{table}[hbt!] 
\caption{\textbf{Ablation results of the design modules of \Pn{} on MPIIGaze-DIR benchmark.}}
    \centering
    \label{table: abl-g}
    \begin{tabular}{l?cccc}
    
        \specialrule{2\heavyrulewidth}{\abovetopsep}{\belowbottomsep}
        \textbf{Metric}&\multicolumn{4}{c}{\textbf{Mean Angle Error (degrees)}$\downarrow$}\\
        \specialrule{1.5\heavyrulewidth}{\abovetopsep}{\belowbottomsep}
        \textbf{Method/Shot} & \textbf{All} & \textbf{Many} & \textbf{Median} & \textbf{Few} \\
        \specialrule{1.5\heavyrulewidth}{\abovetopsep}{\belowbottomsep}
        \textbf{Contrastive-\Pn}&7.11&\textbf{5.00}&7.73&9.89\\
        \textbf{\Pn{}-$\mathcal{S}$}&6.47 & 5.94&6.91&6.71\\

        \textbf{\Pn{}-$Sim$}&6.39&5.69&7.09&6.48\\
        \textbf{\Pn{}-$\eta$}&6.24&5.87&6.96&6.27\\
        \Pn{}~\textbf{(Ours)}&\textbf{6.16}&	5.73&	\textbf{6.85}&	\textbf{6.17}\\
        \specialrule{2\heavyrulewidth}{\abovetopsep}{\belowbottomsep}
        
    \end{tabular}
\end{table}

\paragraph{Similarity Threshold Selection.}
Fig.~\ref{fig: w-in} shows the ablation study on the similarity threshold for IMDB-WIKI-DIR and NYUD2-DIR benchmarks. $\omega = 1$ and $\omega = 5$ are the best similarity threshold choices for IMDB-WIKI-DIR and NYUD2-DIR benchmarks, respectively.

\begin{figure}[hbt!]
  \centering
  
  \begin{subfigure}{0.40\textwidth}    \includegraphics[width=\linewidth]{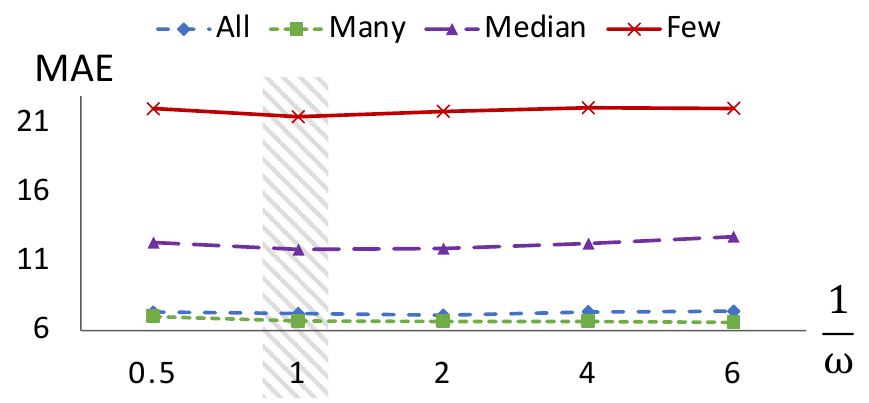}
    \caption{IMDB-WIKI-DIR}
    \label{w-i}
  \end{subfigure}
  \label{fig: abl-w}
  \hfil
  \begin{subfigure}{0.40\textwidth}    \includegraphics[width=\linewidth]{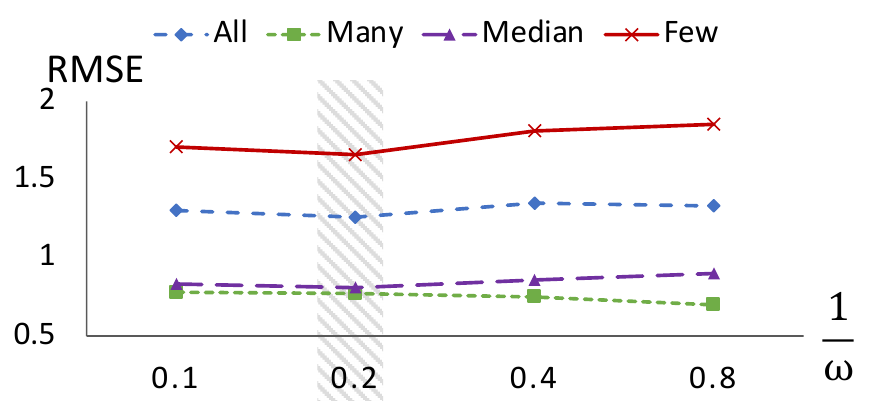}
    \caption{NYUD2-DIR}
    \label{w-n}
  \end{subfigure}
  \begin{subfigure}{0.40\textwidth}    \includegraphics[width=\linewidth]{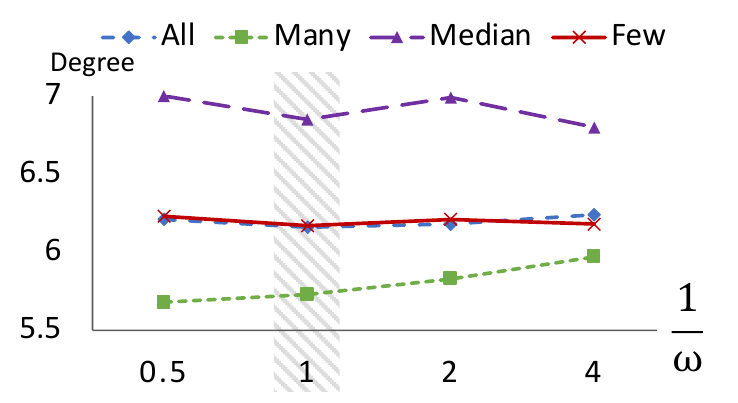}
    \caption{NYUD2-DIR}
    \label{w-g}
  \end{subfigure}
  \caption{ Comparison of different choices of $\omega$ (a) in terms of MAE for IMDB-WIKI-DIR benchmark, (b) in terms of RMSE for NYUD2-DIR and (c) in terms of Mean angular error (degrees) for MPIIGaze-DIR.}
  \label{fig: w-in}
\end{figure}
\paragraph{Hyperparameter selection.} Here we present the selection process of hyperparameters $\alpha$, $\beta$ and $\eta$ for all the benchmarks. Table~\ref{table: hp1-a}, Table~\ref{table: hp1-i}, Table~\ref{table: hp1-n} and Table~\ref{table: hp1-g} show the ablation study of \Pn{} on $\alpha$ and $\beta$ in Eq.~\ref{eq: sum} for AgeDB-DIR, IMDB-WIKI-DIR, NYUD2-DIR and MPIIGaze-DIR benchmarks, respectively. In these tables, hyperparameter $\eta$ is set to the values mentioned in~\ref{implementation details}. Additionally, Table~\ref{table: hp2-a}, Table~\ref{table: hp2-i}, Table~\ref{table: hp2-n} and Table~\ref{table: hp2-g} show the ablation study of \Pn{} on $\eta$ in Eq.~\ref{eq: sim} for AgeDB-DIR, IMDB-WIKI-DIR, NYUD2-DIR and MPIIGaze-DIR benchmarks, respectively. In these tables, hyperparameter $\eta$ is set to the values mentioned in~\ref{implementation details}.
\begin{table}[htb!]
    
    \caption{\textbf{Ablation study on $\alpha$ and $\beta$ in Eq.~\ref{eq: sum} for AgeDB-DIR benchmark}}
    \centering
    \label{table: hp1-a}
    \begin{tabular}{c|c?cccc}
    
        \specialrule{2\heavyrulewidth}{\abovetopsep}{\belowbottomsep}
        \multirow{2}{*}{$\alpha$}&\multirow{2}{*}{$\beta$}&\multicolumn{4}{c}{\textbf{MAE}$\downarrow$}\\
        \cline{3-6}
        
        & & \textbf{All} & \textbf{Many} & \textbf{Median} & \textbf{Few} \\
        \specialrule{2\heavyrulewidth}{\abovetopsep}{\belowbottomsep}
       
        0.5	&1	&7.31	&6.53	&8.08	&10.51\\
        1	&0.5	&7.30	&6.59	&8.10	&10.46\\
        1&	1&	7.31&	6.55	&8.11	&10.48\\
        1&2&7.35&	6.58	&8.07	&10.46\\
\textcolor{green}{\textbf{1}}&	\textcolor{green}{\textbf{4}}	&7.28	&6.53	&8.03	&10.42\\
1	&5	&7.25	&6.47	&8.12	&10.54\\

        \specialrule{2\heavyrulewidth}{\abovetopsep}{\belowbottomsep}
        
    \end{tabular}
\end{table}
\begin{table}[htb!]    
    \caption{\textbf{Ablation study on $\alpha$ and $\beta$ in Eq.~\ref{eq: sum} for IMDB-WIKI-DIR benchmark}}
    \centering
    \label{table: hp1-i}
    \begin{tabular}{c|c?cccc}
    
        \specialrule{2\heavyrulewidth}{\abovetopsep}{\belowbottomsep}
        \multirow{2}{*}{$\alpha$}&\multirow{2}{*}{$\beta$}&\multicolumn{4}{c}{\textbf{MAE}$\downarrow$}\\
        \cline{3-6}
        
        & & \textbf{All} & \textbf{Many} & \textbf{Median} & \textbf{Few} \\
        \specialrule{2\heavyrulewidth}{\abovetopsep}{\belowbottomsep}
       
        0.5	&1	&7.98	&7.11&	14.26&	25.19\\
        1	&0.5	&7.90	&7.17&	14.34&	25.21\\
        1&	1&	7.94	&7.29&	14.28&	25.19\\
        1&2&7.80	&7.21	&14.23&	25.17\\
\textcolor{green}{\textbf{1}}&	\textcolor{green}{\textbf{4}}	&7.84	&7.09	&14.16	&25.15\\
1	&5	&7.98&	7.25&	14.24&	25.13\\

        \specialrule{2\heavyrulewidth}{\abovetopsep}{\belowbottomsep}
        
    \end{tabular}
\end{table}
\begin{table}[hbt!]    
    \caption{\textbf{Ablation study on $\alpha$ and $\beta$ in Eq.~\ref{eq: sum} for NYUD2-DIR benchmark}}
    \centering
    \label{table: hp1-n}
    \begin{tabular}{c|c?cccc}
    
        \specialrule{2\heavyrulewidth}{\abovetopsep}{\belowbottomsep}
        \multirow{2}{*}{$\alpha$}&\multirow{2}{*}{$\beta$}&\multicolumn{4}{c}{\textbf{RMSE}$\downarrow$}\\
        \cline{3-6}
        
        & & \textbf{All} & \textbf{Many} & \textbf{Median} & \textbf{Few} \\
        \specialrule{2\heavyrulewidth}{\abovetopsep}{\belowbottomsep}
        0.5	&1&1.344	&0.722	&0.897&	1.918\\
        1	&0.5&1.324	&0.712	&0.891&	1.945\\
        \textcolor{green}{\textbf{1}}&	\textcolor{green}{\textbf{0.2}} &1.304	&0.682	&0.889	&1.885\\
        1&0.4&1.316	&0.673	&0.909	&1.905\\

        \specialrule{2\heavyrulewidth}{\abovetopsep}{\belowbottomsep}
        
    \end{tabular}
\end{table}

\begin{table}[hbt!]    
    \caption{\textbf{Ablation study on $\alpha$ and $\beta$ in Eq.~\ref{eq: sum} for MPIIGaze-DIR benchmark}}
    \centering
    \label{table: hp1-g}
    \begin{tabular}{c|c?cccc}
    
        \specialrule{2\heavyrulewidth}{\abovetopsep}{\belowbottomsep}
        \multirow{2}{*}{$\alpha$}&\multirow{2}{*}{$\beta$}&\multicolumn{4}{c}{\textbf{Mean Angle Error (degrees)}$\downarrow$}\\
        \cline{3-6}
        
        & & \textbf{All} & \textbf{Many} & \textbf{Median} & \textbf{Few} \\
        \specialrule{2\heavyrulewidth}{\abovetopsep}{\belowbottomsep}
        0.5	&1&6.18&	5.79&	6.89&	6.19\\
        1	&0.2&6.14&	5.85&	6.84&	6.22\\
        \textcolor{green}{\textbf{1}}&	\textcolor{green}{\textbf{0.4}} &6.16&	5.73&	6.85&	6.17\\
        1&1&6.09	&5.64&	7.05&	6.24\\

        \specialrule{2\heavyrulewidth}{\abovetopsep}{\belowbottomsep}
        
    \end{tabular}
\end{table}

\begin{table}[hbt!]    
    \caption{\textbf{Ablation study of $\eta$ (pushing power) for AgeDB-DIR benchmark.}}
    \centering
    \label{table: hp2-a}
    \begin{tabular}{l?cccc}
    
        \specialrule{2\heavyrulewidth}{\abovetopsep}{\belowbottomsep}
        \multirow{2}{*}{$\eta$}&\multicolumn{4}{c}{\textbf{MAE}$\downarrow$}\\
        \cline{2-5}
        
         & \textbf{All} & \textbf{Many} & \textbf{Median} & \textbf{Few} \\
        \specialrule{1.5\heavyrulewidth}{\abovetopsep}{\belowbottomsep}
        0.009&	7.31&	6.63&	7.99&	10.5\\
        \textcolor{green}{\textbf{0.01}}&	7.28	&6.53	&8.03	&10.42\\
        0.05&	7.36&	6.51&	8.11&	10.51\\
        0.1	&7.38	&6.49	&8.09	&10.48\\
        \specialrule{2\heavyrulewidth}{\abovetopsep}{\belowbottomsep}
        
    \end{tabular}
\end{table}

\begin{table}[hbt!]    
    \caption{\textbf{Ablation study of $\eta$ (pushing power) for IMDB-WIKI-DIR benchmark.}}
    \centering
    \label{table: hp2-i}
    \begin{tabular}{l?cccc}
    
        \specialrule{2\heavyrulewidth}{\abovetopsep}{\belowbottomsep}
        \multirow{2}{*}{$\eta$}&\multicolumn{4}{c}{\textbf{MAE}$\downarrow$}\\
        \cline{2-5}
        
         & \textbf{All} & \textbf{Many} & \textbf{Median} & \textbf{Few} \\
        \specialrule{1.5\heavyrulewidth}{\abovetopsep}{\belowbottomsep}
        0.009&	7.88	&7.07	&14.25	&25.16\\
        \textcolor{green}{\textbf{0.01}}&	7.84 &7.09 &14.16 &25.15\\
        0.05&	7.88&	6.99&	14.20&	25.16\\
        0.1	&7.91	&7.14&	14.18&	25.21\\
        \specialrule{2\heavyrulewidth}{\abovetopsep}{\belowbottomsep}
        
    \end{tabular}
\end{table}

\begin{table}[hbt!]    
    \caption{\textbf{Ablation study of $\eta$ (pushing power) for NYUD2-DIR benchmark.}}
    \centering
    \label{table: hp2-n}
    \begin{tabular}{l?cccc}
    
        \specialrule{2\heavyrulewidth}{\abovetopsep}{\belowbottomsep}
        \multirow{2}{*}{$\eta$}&\multicolumn{4}{c}{\textbf{RMSE}$\downarrow$}\\
        \cline{2-5}
        
         & \textbf{All} & \textbf{Many} & \textbf{Median} & \textbf{Few} \\
        \specialrule{1.5\heavyrulewidth}{\abovetopsep}{\belowbottomsep}
        0.1	&1.312&	0.674&	0.886&	1.891\\
        \textcolor{green}{\textbf{0.2}}	&1.304	&0.682&	0.889&	1.885\\
        0.4	&1.295	&0.619&	0.92&	1.911\\

        \specialrule{2\heavyrulewidth}{\abovetopsep}{\belowbottomsep}
        
    \end{tabular}
\end{table}

\begin{table}[hbt!]    
    \caption{\textbf{Ablation study of $\eta$ (pushing power) for MPIIGaze-DIR benchmark.}}
    \centering
    \label{table: hp2-g}
    \begin{tabular}{l?cccc}
    
        \specialrule{2\heavyrulewidth}{\abovetopsep}{\belowbottomsep}
        \multirow{2}{*}{$\eta$}&\multicolumn{4}{c}{\textbf{Mean angular error (degrees)}$\downarrow$}\\
        \cline{2-5}
        
         & \textbf{All} & \textbf{Many} & \textbf{Median} & \textbf{Few} \\
        \specialrule{1.5\heavyrulewidth}{\abovetopsep}{\belowbottomsep}
        0.5	&6.15	&5.77	&6.83	&6.21\\
        \textcolor{green}{\textbf{1}}	&6.16	&5.73	&6.85	&6.17\\
        2	&6.18&	5.74	&6.89	&6.23\\

        \specialrule{2\heavyrulewidth}{\abovetopsep}{\belowbottomsep}
        
    \end{tabular}
\end{table}

\paragraph{Contrastive Regression.}
To evaluate the impact of contrastive learning on deep imbalanced regression, we use two contrastive regularizers: \textbf{MoCo}: We regularize the baselines with infoNCE loss, using the architecture of with MoCo V1~\citep{moco}, MoCo V2~\citep{chen2020improved}, and \Pn{}. Here we regularized a regression model with both Moco v1 and MoCo v2. Our experiments shows that MoCo v2 degrades the regression performance in some cases.
Table~\ref{table: CON} compares the performance of a regression model on AgeDB-DIR, IMDB-WIKI-DIR and NYUD2-DIR benchmarks when it is regularized in a contrastive manner with MoCo V1, MoCo V2,  and \Pn. The results are reported in terms of MAE for AgeDB-DIR benchmark and IMDB-WIKI-DIR benchmark and in terms of RMSE for NYUD2-DIR dataset.
Moco considerably boost the performance of  VANILLA and shows that contrastive training significantly improves the regression performance, especially for minority samples. 
\Pn{} incorporate unbiased supervision into the contrastive regression and significantly boost the performance on minority samples with no harm to the learning process for majority samples. 
As shown in Fig.~\ref{fig: cont}, Moco and \Pn{} provide more consistent performance compared to the baseline. In addition, Moco is consistently outperformed by \Pn{} and empirically confirms \Pn{} improves the self-supervised contrastive regularizer by incorporating supervision in an unbiased manner. 
\begin{table}[hbt!]
\caption{\textbf{Results of contrastive learning analysis on AgeDB-DIR, IMDB-WIKI-DIR and NYUD2-DIR benchmarks.} Results are reported for the whole test data (\textit{all}) and three other shots: \textit{many}, \textit{median}, \textit{few}. At the bottom of the table  the improvements of \Pn{} with respect to Moco are reported in \textcolor{green}{green} for each benchmark, shot and metric. In each column, the best result is in \textbf{bold}. }
    \resizebox{\textwidth}{!}{%
    \centering
    
    \label{table: CON}
    \begin{tabular}{l?cccc:cccc:cccc}
    
        \specialrule{2\heavyrulewidth}{\abovetopsep}{\belowbottomsep}
        \textbf{Benchmark}&\multicolumn{4}{c:}{\textbf{AgeDB-DIR} }&\multicolumn{4}{c:}{\textbf{IMDB-WIKI-DIR} }&\multicolumn{4}{c}{\textbf{NYUD2-DIR} }\\
        \specialrule{1\heavyrulewidth}{\abovetopsep}{\belowbottomsep}
        \textbf{Metric}&\multicolumn{4}{c:}{\textbf{MAE}$\downarrow$}&\multicolumn{4}{c:}{\textbf{MAE}$\downarrow$}&\multicolumn{4}{c}{\textbf{RMSE}$\downarrow$}\\
        \specialrule{.8\heavyrulewidth}{\abovetopsep}{\belowbottomsep}
        \backslashbox{\textbf{Method}}{\textbf{Shot}} & \textbf{All} & \textbf{Many} & \textbf{Median} & \textbf{Few}& \textbf{All} & \textbf{Many} & \textbf{Median} & \textbf{Few}& \textbf{All} & \textbf{Many} & \textbf{Median} & \textbf{Few} \\
       \specialrule{.8\heavyrulewidth}{\abovetopsep}{\belowbottomsep}
        VANILLA &7.35&6.56&8.23&12.37&8.06&7.23&15.12&26.33&1.477 &\textbf{0.591} &0.952 &2.123\\ 
               
        + Moco V1&7.33&\textbf{6.50}&8.19&11.72& 7.89 &7.13&14.78&26.11&1.370&0.601&0.902&1.912\\
        + Moco V2&7.47&6.21&8.75&12.75& 8.12 &6.99&15.02&26.01&1.404&0.632&0.978&2.207\\
        + \Pn &\textbf{7.28}&6.53&\textbf{8.03}&\textbf{10.42}&\textbf{7.84} &\textbf{7.09} &\textbf{14.16} &\textbf{25.1}5&\textbf{1.304}&0.682&\textbf{0.889}&\textbf{1.885}\\
        
        \hline\hline
        \Pn{} \textit{vs.} Moco & \textcolor{green}{0.68\%}&\textcolor{blue}{-0.46\%}&\textcolor{green}{1.96\%}&\textcolor{green}{11.10\%}&\textcolor{green}{0.63\%}&\textcolor{green}{0.56\%}&\textcolor{green}{4.20\%}&\textcolor{green}{3.68\%}&\textcolor{green}{4.82\%}&\textcolor{blue}{-1.63\%}&\textcolor{green}{1.44\%}&\textcolor{green}{4.41\%}\\
        
        \specialrule{2\heavyrulewidth}{\abovetopsep}{\belowbottomsep}
    \end{tabular}
    }
\end{table}

\begin{figure}[hbt!]
  \centering
  \begin{subfigure}{0.32\textwidth}
    \includegraphics[width=1\linewidth,height =32mm]{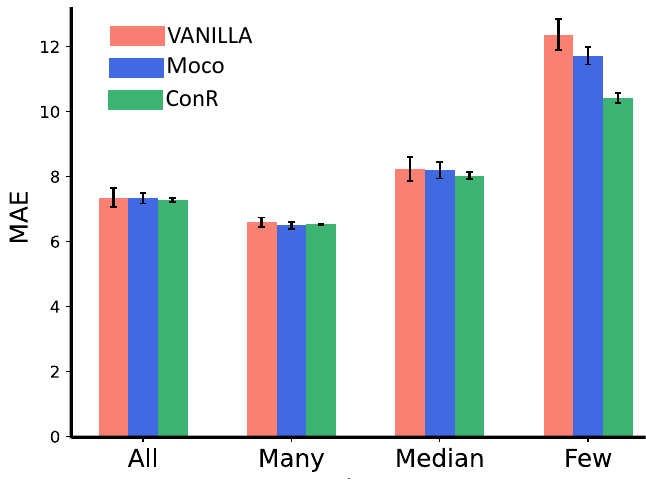}
    \caption{AgeDB-DIR}
  \end{subfigure}
  \hfill
  \begin{subfigure}{0.32\textwidth}   \includegraphics[width=1\linewidth,height =32mm]{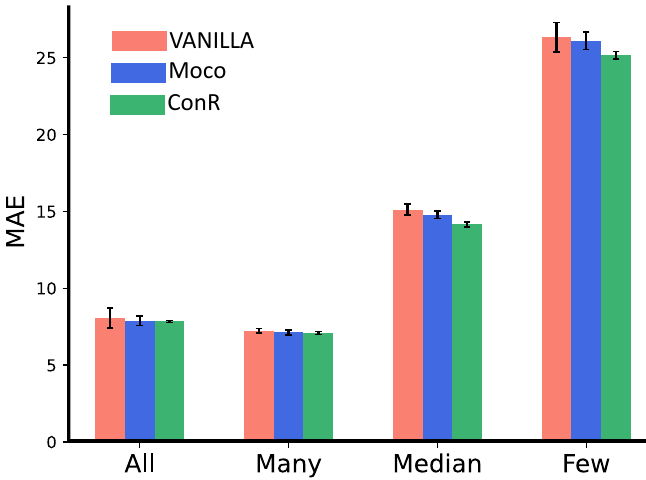}
    \caption{IMDB-WIKI-DIR}
  \end{subfigure}
  \hfill
  \begin{subfigure}{0.32\textwidth}  \includegraphics[width=1\linewidth,height =32mm]{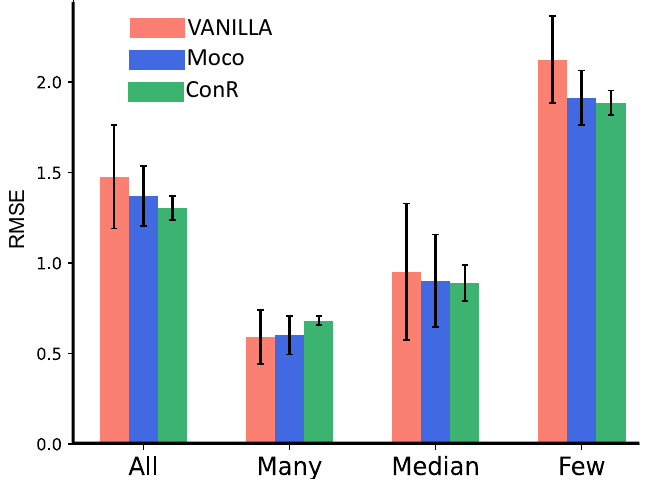}
    \caption{NYUD2-DIR}
  \end{subfigure}
  \caption{Comparison of the performance gain of regularizing VANILLA regression model with  contrastive regularizers: Moco V1 and \Pn{}  on (a)~AgeDB-DIR, (b)~IMDB-WIKI-DIR anf (c)~NYUD2-DIR benchmarks. It shows contrastive regularizer consistently lifts the performance of the baseline, particularly on minority samples. }
  \label{fig: cont}
\end{figure}
\section{More Theoretical Insights}
\label{appendix:theoretical}
Here we theoretically justify the effectiveness of \Pn{} by deriving a upper bound on the probability of incorrect labelling of minority samples. We show that minimizing $\mathcal{L}_{ConR}$ robustly minimizes this probability of mislabeling for minority samples and consequently improves the generalizability.

In the following, we'll derive the upper bound: following \citep{oord2018representation, wang2022contrastive} we define the density ratio $f(\hat y,x)$ in InfoNCE~\cite{oord2018representation} to be $\frac{p(\hat y|x)}{p(\hat y)}$ \citep{oord2018representation, wang2022contrastive} that is estimated by $\exp(z_j\cdot z_i/\tau)$ in Eq.~\ref{eq: conr} like other contrastive objective functions~\citep{moco, pmlr-v119-chen20j}. $p(\hat y|x)$ is the desired prediction distribution and $\hat Y_j = (\hat y_j -\omega, \hat y_j + \omega)$. 

For each anchor $x_j$ the ${\mathcal{L}_{ConR}}_j$ in Eq.~\ref{eq: conr} is defined to be:

\begin{align}
    {\mathcal{L}_{ConR}}_j = -\log \frac{1}{N^+_j} 
         \sum_{z_i \in K^+_j} \frac{
             \exp(z_j\cdot z_i/\tau)
         }{\sum_{z_p \in K^+_j} 
                 \exp(z_j\cdot z_p/\tau) 
             + \sum_{z_q \in K^-_j}
                \mathcal{S}_{j,q} \exp(z_j\cdot z_q/\tau)
        }
\end{align}

Then, ${\mathcal{L}_{ConR}}_j$ can be rewritten as:
\begin{align}
     {\mathcal{L}_{ConR}}_j
         =& -\log \frac{1}{N^+_j}
         \sum^{ N^+_j}_{i=0} \frac{
             f(\hat y_j,x_i)
         }{\sum^{K^+_j}_{p=0} 
                 f(\hat y_j,x_p) 
             + \sum^{K^-_j}_{q=0}
                \mathcal{S}_{j,q} f(\hat Y_j,x_q)
        }
\end{align}

Following Jensen's inequality we have:
\begin{align}
   {\mathcal{L}_{ConR}}_j =& -\log \frac{1}{N^+_j}
         \sum^{N^+_j}_{i=0} \frac{
             f(\hat y_j,x_i)
         }{\sum^{N^+_j}_{p=0} 
                 f(\hat y_j,x_p) 
             + \sum^{K^-_j}_{q=0}
                \mathcal{S}_{j,q} f(\hat Y_j,x_q)
        }\\
        \overset{\text{\cite{mcshane1937jensen}}}{\ge}& - \frac{1}{N^+_j}
         \sum^{N^+_j}_{i=0} \log \frac{
             f(\hat y_j,x_i)
         }{\sum^{K^+_j}_{p=0} 
                 f(\hat y_j,x_p) 
             + \sum^{N^-_j}_{q=0}
                \mathcal{S}_{j,q} f(\hat Y_j,x_q)
        }
\end{align}

where $x_q$ are the negative sample, and $\hat y_j$ is the prediction of positive samples. prediction of $x_q$ mistakenly fall in the range of $(\hat y_j \pm \omega)$.
Further, we have:
\begin{align}
     {\mathcal{L}_{ConR}}_j  
        \ge& - \frac{1}{N^+_j}
         \sum^{K^+_j}_{i=0} \log \frac{
             f(\hat y_j,x_i)
         }{\sum^{N^+_j}_{p=0} 
                 f(\hat y_j,x_p) 
             + \sum^{N^-_j}_{q=0}
                \mathcal{S}_{j,q} f(\hat Y_j,x_q)
        }\\
        =& \frac{1}{N^+_j}\sum^{N^+_j}_{i=0} \log \Biggl[\frac{\sum^{ N^+_j}_{p=0} 
                 f(\hat y_j,x_p)}{f(\hat y_j,x_i)} + \frac{\sum^{ N^-_j}_{q=0}
                \mathcal{S}_{j,q} f(\hat Y_j,x_q)}{f(\hat y_j,x_i)}\Biggr]\\
        \ge& \frac{1}{N^+_j}\sum^{ N^+_j}_{i=0} \log \Biggl[\frac{\sum^{ N^+_j}_{p=0} 
                 f(\hat y_j,x_p)}{\sum^{ N^+_j}_{i=0} 
                 f(\hat y_j,x_i)} + \frac{\sum^{ K^-_j}_{q=0}
                \mathcal{S}_{j,q} f(\hat Y_j,x_q)}{f(\hat y_j,x_i)}\Biggr]\\
        \ge& \frac{1}{N^+_j}\sum^{ N^+_j}_{i=0} \log \Biggl[1 + \frac{\sum^{N^-_j}_{q=0}
                \mathcal{S}_{j,q} f(\hat Y_j,x_q)}{f(\hat y_j,x_i)}\Biggr]\\
        \ge& \frac{1}{N^+_j}\sum^{ N^+_j}_{i=0} \log \Biggl[ \frac{\sum^{N^-_j}_{q=0}
                \mathcal{S}_{j,q} f(\hat Y_j,x_q)}{f(\hat y_j,x_i)}\Biggr]\\
        =& \frac{1}{N^+_j} \big[\log \sum^{N^-_j}_{q=0}
                \mathcal{S}_{j,q} f(\hat Y_j,x_q) -  \log \sum^{ N^+_j}_{i=0} f(\hat y_j,x_i)\big]
\end{align}
Next, the $ {\mathcal{L}_{ConR}}$ is:
\begin{align}
     {\mathcal{L}_{ConR}} =& \frac{1}{2N}\sum^{2N}_{j=0}{\mathcal{L}_{ConR}}_j \\ 
        \ge& \frac{1}{2N}\sum^{2N}_{j=0}\frac{1}{N^+_j} \big[\log \sum^{N^-_j}_{q=0}
                \mathcal{S}_{j,q} f(\hat Y_j,x_q) -  \log \sum^{ N^+_j}_{i=0} f(\hat y_j,x_i)\big]\\
        \ge& \frac{1}{2N}\sum^{2N}_{j=0}\frac{1}{2N} \big[\log \sum^{N^-_j}_{q=0}
                \mathcal{S}_{j,q} f(\hat Y_j,x_q) -  \log \sum^{ N^+_j}_{i=0} f(\hat y_j,x_i)\big]\\
        =& \frac{1}{4N^2}\sum^{2N}_{j=0} \big[\log \sum^{N^-_j}_{q=0}
                \mathcal{S}_{j,q} f(\hat Y_j,x_q) -  \log \sum^{ N^+_j}_{i=0} f(\hat y_j,x_i)\big]\\
        =& \frac{1}{4N^2}(\sum^{2N}_{j=0} \log \sum^{N^-_j}_{q=0}
                \mathcal{S}_{j,q} f(\hat Y_j,x_q)- \sum^{2N}_{j=0}  \log \sum^{ N^+_j}_{i=0} f(\hat y_j,x_i))\\
        =& \frac{1}{4N^2}(\sum^{2N}_{j=0} \log \sum^{N^-_j}_{q=0}
                \mathcal{S}_{j,q} f(\hat Y_j,x_q)- \sum^{2N}_{j=0}  \log \sum^{ N^+_j}_{i=0} \frac{p(\hat y_j| x_i)}{p(\hat y_j)})\\
        \overset{\frac{p(\hat y_j|x_i)}{p(\hat y_j)}<1}{\ge}& \frac{1}{4N^2}(\sum^{2N}_{j=0} \log \sum^{N^-_j}_{q=0}
                \mathcal{S}_{j,q} f(\hat Y_j,x_q)- \sum^{2N}_{j=0}  \log \sum^{ N^+_j}_{i=0} 1)\\
        =& \frac{1}{4N^2}(\sum^{2N}_{j=0} \log \sum^{N^-_j}_{q=0}
                \mathcal{S}_{j,q} f(\hat Y_j,x_q)- 2N\log N^+_j).
\end{align}

Further, we have:
\begin{align}
        {\mathcal{L}_{ConR}}=&\frac{1}{4N^2}(\sum^{2N}_{j=0} \log \sum^{N^-_j}_{q=0}
                \mathcal{S}_{j,q} f(\hat Y_j,x_q)- 2N\log N^+_j)\\
        \ge&\frac{1}{4N^2}(\sum^{2N}_{j=0} \sum^{N^-_j}_{q=0}  \log
                \mathcal{S}_{j,q} f(\hat Y_j,x_q)- 2N\log N^+_j)\\
        =&\frac{1}{4N^2}(\sum^{2N}_{j=0} \sum^{N^-_j}_{q=0}  \log
                \mathcal{S}_{j,q} \frac{p(\hat Y_j|x_q)}{p(\hat Y_j)}- 2N\log N^+_j)\\
        \overset{p(\hat Y_j)<1}{\ge}& \frac{1}{4N^2}(\sum^{2N}_{j=0} \sum^{N^-_j}_{q=0}  \log
                \mathcal{S}_{j,q} p(\hat Y_j|x_q) - 2N\log N^+_j)\\
        \ge& \frac{1}{4N^2}(\sum^{2N}_{j=0} \sum^{N^-_j}_{q=0}  \log
                \mathcal{S}_{j,q} p(\hat Y_j|x_q) - 2N\log (2N)).
\end{align}

Then:

\begin{align}
    \frac{1}{4N^2}\sum^{2N}_{j=0} \sum^{ N^-_j}_{q=0}\log\mathcal{S}_{j,q}
                 p(\hat Y_j|x_q) \le \mathcal{L}_{ConR} + \frac{log(2N)}{2N}.      
\end{align}

As explained in \ref{method}, among the 2N augmented samples only the ones with the confusion around them are chosen as anchors. Assuming $A$ is the set of selected anchors that is a subset of the 2N augmented samples we have:
\begin{align}
\label{bound}
    \frac{1}{4N^2}\sum^{2N}_{j=0, x_j \in A} \sum^{ K^-_j}_{q=0}\log\mathcal{S}_{j,q}
                 p(\hat Y_j|x_q)  \le \mathcal{L}_{ConR} + \epsilon, \;\; \epsilon \overset{N\rightarrow \infty}{\rightarrow}  0   
\end{align}
In Eq.~\ref{bound}, $p(\hat Y_j|x_q)$ is the likelihood of sample $x_q$ with an incorrect prediction $y_q \in \hat Y_j$. We refer to $p(\hat Y_j|x_q)$ as the probability of collapse for $x_q$.
The left-hand side presents this probability for all the negative pairs. 

Regarding the empirical study by ~\citet{yang2021delving}, when learning a regression function from the imbalanced data, the representations of minority samples tend to collapse to the majority ones. Since in the definition of ${\mathcal{L}_{ConR}}_j$ in Eq.~\ref{eq: conr}, $x_q$ show the collapsed minority samples, minimizing the left side of the inequality in Eq.~\ref{bound} is the intended optimization in deep imbalanced regression.

The left-hand side is the probability of all collapses during the training and regarding Eq.~\ref{bound} Convergence of $\mathcal{L}_{ConR}$ tightens the upper bound for it.
Minimizing the left-hand side can be explained with two non-disjoint scenarios: either the number of anchors or the degree of collapses is reduced. Here the degree of collapse for each negative sample refers to the quantified disagreement between the label similarity and prediction similarity as discussed in section ~\ref{method}.
In addition, each collapse probability is weighted with $\mathcal{S}_{j,q}$, leading to penalizing the incorrect predictions with regard to their severity.
In other words, \Pn{} penalizes the bias probabilities with a re-weighting scheme, where the weights are defined based on the agreement of the predictions and labels.

\section{Algorithm of \Pn}

Algorithm~\ref{alg:conr} shows the pseudo-code of regularizing a regression model using \Pn{}.
In this algorithm, $\mathcal{S}_{j,q}$ is the pushing weights for selected anchor $z_j$ and its negative pair $z_q$~(Eq.~\ref{eq: sim}). $K^+_j$ and $K^-_j$ are positive pairs of $z_j$, and negative pairs of $z_j$, respectively.

\begin{algorithm}
\caption{\Pn: Contrastive regularizer for deep imbalanced regression}\label{alg:conr}
\begin{algorithmic}
\Require \\\quad input samples $\{(x_i,y_i)\}^{N}_{i=0}$ \\\quad feature encoder $\mathcal{E}(.)$ \\\quad regression function $\mathcal{R}(.)$
\For {e in epochs}
\For{ $b = \{(x_i,y_i)\}^{N_b}_{i=0}$ in Batches }
\State $\{x^a_j,y_j\}^{2N_b}_{j=0} \overset{\text{Augmentations}}{\leftarrow}   \{x_i,y_i\}^{N_b}_{i=0} $
\State $\{z_j\}^{2N}_{j=0} \leftarrow \mathcal{E}(\{x^a_j\}^{2N_b}_{j=0})$\;
\Comment{$z_j$ : feature embeddings of augmented input $x^a_j$. }
\State $\{\hat y_j\}^{2N_b}_{j=0} \leftarrow \mathcal{R}(\mathcal{E}(\{z_j\}^{2N_b}_{j=0})) $
\Comment{$\hat y_j$ : prediction of  $z_j$. }
\State $\{K^+_j,K^-_j\}^{2N_b}_{j=0} \leftarrow \{ z_j,y_j, \hat y_j \}^{2N_b}_{j=0}$
\For{$ j \in (0,2N_b)$}
\If {$N^-_j == 0$}
\State ${\mathcal{L}_{ConR}}_j \leftarrow 0$
\Comment{Samples with no negative pairs are not selected as anchors.}
\Else{}
\State  $\{\mathcal{S}_{j,q}\}^{N^-_j}_{q=0} \leftarrow f_{\mathcal{S}}(y_j,{\{y_q\}}^{N^-_j}_{q=0})$
\Comment $\mathcal{S}_{j,q}$: pushing weight~(Eq.~\ref{eq: sim})
\State ${\mathcal{L}_{ConR}}_j \leftarrow \text{computeConR}(z_j,K^+_j,K^-_j,{\{\mathcal{S}_{j,q}\}}^{N^-_j}_{q=0})$
\Comment{Eq.~\ref{eq: conr}}
\EndIf
\EndFor
\State $\mathcal{L}_{ConR} \overset{\text{Average}}
{\leftarrow} {\{{\mathcal{L}_{ConR}}_j\}}^{2N_b}_{j=0}$
\Comment{Eq.~\ref{eq:conr-all}}
\State $\mathcal{L}_\mathcal{R} \leftarrow \text{computeRegressionLoss}({\{\hat y_j, y_j\}}^{N_b}_{j=0})$
\State $L_{sum} \leftarrow \alpha \mathcal{L}_\mathcal{R} + \beta \mathcal{L}_{ConR}$
\State $L_{sum}.\text{backPropagate()}$
\EndFor
\EndFor
\Ensure Trained regression function $\mathcal{R}(.)$ and trained feature encoder $\mathcal{E}(.)$

\end{algorithmic}
\end{algorithm}


    
    
 
 


\end{document}